\documentclass[sigconf, nonacm]{acmart}

\usepackage[normalem]{ulem}
\usepackage{graphicx}
\usepackage{balance}  
\usepackage{multirow}
\usepackage{tabularx}
\usepackage{hyperref}
\usepackage{graphicx}
\usepackage{xspace}
\usepackage{caption}
\DeclareCaptionType{copyrightbox}
\usepackage{subcaption}
\usepackage{color}
\usepackage{tikz}
\usepackage{listings}
\usepackage{epstopdf}
\usepackage[lined,ruled,linesnumbered]{algorithm2e}
\usepackage{booktabs}
\usepackage{subcaption}
\usepackage{lscape}
\usepackage{lipsum}
\usepackage{multicol}
\usepackage{pifont}
\usepackage{soul}
\usepackage{caption}
\usepackage{svg}
\usepackage{longtable}
\usepackage{booktabs}
\usepackage{array}
\usepackage{caption}
\usepackage{rotating}

\usepackage{array}
\newcolumntype{L}[1]{>{\raggedright\let\newline\\\arraybackslash\hspace{0pt}}m{#1}}
\newcolumntype{C}[1]{>{\centering\let\newline\\\arraybackslash\hspace{0pt}}m{#1}}
\newcolumntype{R}[1]{>{\raggedleft\let\newline\\\arraybackslash\hspace{0pt}}m{#1}}

\newcommand{\cmark}{\ding{51}}%
\newcommand{\xmark}{\ding{55}}%
\newcommand{\squishlist}{
  \begin{list}{$\bullet$}
   {
     \setlength{\itemsep}{0pt}
     \setlength{\parsep}{0pt}
     \setlength{\topsep}{0pt}
     \setlength{\partopsep}{0pt}
     \setlength{\leftmargin}{1.5em}
     \setlength{\labelwidth}{1em}
     \setlength{\labelsep}{0.5em} } }
\newcommand{\squishend}{
   \end{list}  }

\definecolor{annotationColor}{RGB}{0,0,139}

\newtheorem{exmp}{Example}

\newcommand{\system}{CBench\xspace}

\newcommand\vldbdoi{10.14778/3457390.3457398}

\begin{document}
\title{\system: Towards Better Evaluation of Question Answering Over Knowledge Graphs}

 \author{Abdelghny Orogat}
 \affiliation{%
   \institution{Carleton University}
 }
 \email{abdelghny.orogat@carleton.ca}

 \author{Isabelle Liu}
 \affiliation{%
   \institution{Carleton University}
 }
  \email{isabelle.liu@carleton.ca}

 \author{Ahmed El-Roby}
 \affiliation{%
   \institution{Carleton University}
 }
\email{ahmed.elroby@carleton.ca}


\begin{abstract}
 Recently, there has been an increase in the number of knowledge graphs that can be only queried by experts. However, describing questions using structured queries is not straightforward for non-expert users who need to have sufficient knowledge about both the vocabulary and the structure of the queried knowledge graph, as well as the syntax of the structured query language used to describe the user's information needs. The most popular approach introduced to overcome the aforementioned challenges is to use natural language to query these knowledge graphs. Although several question answering benchmarks can be used to evaluate question-answering systems over a number of popular knowledge graphs, choosing a benchmark to accurately assess the quality of a question answering system is a challenging task.
 
 \textcolor{black}{
 In this paper, we introduce \system, an extensible, and more informative benchmarking suite for analyzing benchmarks and evaluating question answering systems. \system can be used to analyze existing benchmarks with respect to several fine-grained linguistic, syntactic, and structural properties of the questions and queries in the benchmark. We show that existing benchmarks vary significantly with respect to these properties deeming choosing a small subset of them unreliable in evaluating QA systems. Until further research improves the quality and comprehensiveness of benchmarks, \system can be used to facilitate this evaluation using a set of popular benchmarks that can be augmented with other user-provided benchmarks. \system not only evaluates a question answering system based on popular single-number metrics but also gives a detailed analysis of the linguistic, syntactic, and structural properties of answered and unanswered questions to better help the developers of question answering systems to better understand where their system excels and where it struggles.}

\end{abstract}

\maketitle



\section{Introduction}
\label{sec:introduction}

Recent years witnessed an unprecedented growth in the number of knowledge graphs (KGs)~\cite{Auer2007,Bollacker2008,carlson2010toward,matuszek2006introduction,Suchanek2007,vrandevcic2014wikidata}. 
These knowledge graphs contain a plethora of information that can be potentially used for question answering (QA). However, finding answers in a KG is not an easy task. The user is required to have a detailed knowledge of the KG, and a structured query language to describe their questions in a structured format that can be used to find matches in the KG. This requirement limits the ability to ask questions to power users who have the necessary skills to write syntactically and semantically correct queries to accurately represent their information needs. The number of such users represents a tiny fraction of a potentially large userbase. To overcome this challenge, a large number of QA systems that let users describe their information needs using natural language were developed~\cite{Cabrio2012, Cui2017, Dong2007, Fader2014, Kaufmann2006, Lopez2011, Unger2012, Yahya2013, Zenz2009, Zheng2018, Zhou2007}. In fact, over 62 QA systems have been developed since 2010~\cite{Hoeffner2017}. 

As a result of the popularity of QA over KGs, several benchmarks were introduced to evaluate QA systems~\cite{Azmy2018,Berant2013,Cabrio2013,Su2016,trivedi2017lc,Unger2011,Unger2014,Unger2015,Unger2016,Usbeck2017}. These benchmarks typically include questions described in natural language, answers to the questions from the KG targeted by the benchmark, and possibly structured queries that return the previously mentioned answers. To evaluate a newly developed QA system, its developers need to choose from a large number of benchmarks \textcolor{black}{(at least 17 at the time of writing this paper)} to evaluate their system. Without a quantitative comparison that highlights the differences between these benchmarks, choosing a subset of them to evaluate a new QA system is mainly motivated by the ease of comparison to existing systems in the literature
\footnote{\textcolor{black}{Out of 20 QA systems that were developed in the past 5 years, only 6 are open-sourced, and only 3 systems are accessible via a functional web interface.}}
rather than by how effective a benchmark is in evaluating a QA system. In fact, existing benchmarks differ significantly from each other with respect to the following three points: 
\squishlist
\item How the benchmarks were created: Some benchmarks are manually created by human experts based on heuristics \cite{abujabal-etal-2019-comqa, bordes2015large, Unger2011, cabrio2012qakis, Cabrio2013, Unger2014, Unger2015, Unger2016, Usbeck2017, Usbeck2018,  ngomo20189th}. Other benchmarks are automatically generated from the KG~\cite{Su2016,trivedi2017lc,Azmy2018}.
\item \textcolor{black}{Metadata}: This includes what KGs that the benchmark target, and the number of questions in the benchmark. Most benchmarks target a limited number of KGs (All benchmarks that we are aware of target only 5 KGs). Also, the number of questions in each benchmark varies significantly. Out of the benchmarks we are aware of, the benchmark with the smallest number of questions includes 150 questions, while the benchmark with the largest number of questions includes 108,442 questions.
\item \textcolor{black}{Linguistic, syntactical, and structural properties: The natural language questions have linguistic properties. Their corresponding queries (if they exist in the benchmark) also have syntactical and structural properties. In this paper, we reveal that existing benchmarks vary significantly with respect to these properties, which is reflected on the reported quality scores depending on the benchmark used.} 
\squishend
Another issue with existing benchmarks is their limited usability. Today, benchmarks are used in the following fashion: The user parses the benchmark file, extract the questions and utilizes the QA system to find answers in the targeted KG, then compare the returned answers to the answers extracted from the benchmark file to calculate multiple evaluation scores like micro, macro, and global F-1 scores (discussed in Section~\ref{sec:eval}). The user then examines the questions that the QA system failed to answer correctly and debug their code to identify why the QA system struggles with these questions. In that sense, the benchmark is used as a dataset that helps in producing the aforementioned scores with a lost potential of being more informative to its user by giving more details on the fine-grained properties of the processed questions, which will help the user better understand how the QA system behaves.

\textcolor{black}{In this paper, we introduce \system\footnote{\url{https://github.com/aorogat/CBench}}, an extensible fine-grained benchmark suite that overcomes the aforementioned challenges facing accurate evaluation of QA systems. \system can be used in two modes: \emph{Benchmark Analysis Mode} and \emph{QA Evaluation Mode}. \system can be used in the first mode to perform a fine-grained analysis on the natural language questions and queries in a set of benchmarks selected by the user. \system includes a total of 17 benchmarks targeting 5 KGs, and can be easily extended with new benchmarks and KGs. For the structured queries, \system analyzes several syntactical properties of structured queries like the type of the query, the operators used, and the query size. \system also analyzes the structural properties (shapes) of the queries. For the natural language questions, \system analyzes several linguistic properties like the type of question, and part-of-speech (PoS) tags of each question token of the natural language question. In our analysis, we surprisingly quantify high-degree variations with respect to the linguistic, syntactical, and structural properties of questions and queries among different benchmarks in the literature, and experimentally show that these differences result in an inconsistent assessment of QA systems. These findings motivate further research in building better benchmarks that address as much fine-grained properties as possible to have a good coverage of real-world questions that the QA systems will encounter in real-world deployments.}

\textcolor{black}{Until such benchmarks are available, CBench facilitates evaluating QA systems in the second mode (\emph{QA Evaluation Mode}), in which the user can choose the benchmarks they wish to use to evaluate their QA system. Then, \system interacts with the QA system using a set of well-defined APIs to send/recieve questions/answers. In addition to reporting micro, macro, and global F-1 scores, \system also analyzes all the questions in the chosen benchmarks and their corresponding structured queries (if available). Specifically, \system returns (1) a detailed analysis of the properties of the queries that the evaluated QA system processed, and (2) linguistically-similar natural language questions for any question of interest (e.g., a question that the QA system failed to answer). Using the two aforementioned types of output, the QA user can either (1) identify common properties between questions that the QA system struggles with (e.g., most of the questions have a specific query shape), or (2) identify obvious inconsistencies in the processed questions. For example, using \system, we were able to quickly identify that one of the QA systems we evaluated was able to answer the question ``What is the capital of Cameroon?'' correctly, while it incorrectly answered ``What is the capital of Canada?'', which highlights an overfitting problem in their entity recognition and relation mapping approaches. Being able to quickly identify commonalities or inconsistencies will help the QA system developers to quickly identify the QA system component that they need to improve. Based on the insights provided by \system, the user can also use it in a \emph{Debugging Mode} within the \emph{QA Evaluation Mode}, where they can control \system's output questions based on any of the linguistic, syntactical, or structural properties of all the questions and queries in \system to better understand how their QA system behaves in several controlled situations. For example, the user can specify that \system only outputs temporal questions whose queries have a star-shape.}

Our contributions in the \system suite are:
\textcolor{black}{
\squishlist
 \item To the best of our knowledge, We are the first to introduce the concept of fine-grained analysis of questions and queries in QA benchmarks. 
 \item Using our fine-grained analysis, we surprisingly identify a high degree of variations among existing benchmarks with respect to several linguistic, syntactical, and structural properties of the natural language questions and their corresponding queries in the benchmarks.
 \item To demonstrate the effects of such variations, we evaluate six QA systems using \system in the \emph{QA Evaluation Mode} and show that their quality scores vary significantly.
 \item We give important insights to QA systems' researchers on benchmark selection and QA evaluation metrics.
\squishend
}
The rest of this paper is organized as follows: 
Section~\ref{sec:preliminaries} presents the preliminaries of this paper. 
Section~\ref{sec:architecture} discusses the architecture of \system. 
Section~\ref{sec:benchmarks} gives an overview of the benchmarks in \system. 
Section~\ref{sec:structuredAnalysis} presents how \system analyzes the structured queries in the benchmarks. 
Section~\ref{sec:nl} presents how \system analyzes the natural language questions. 
Section~\ref{sec:eval} discusses our experiments on six QA systems using \system. 
Section~\ref{sec:conclusion} concludes the paper.

\usetikzlibrary {positioning}

\section{Preliminaries}
\label{sec:preliminaries}

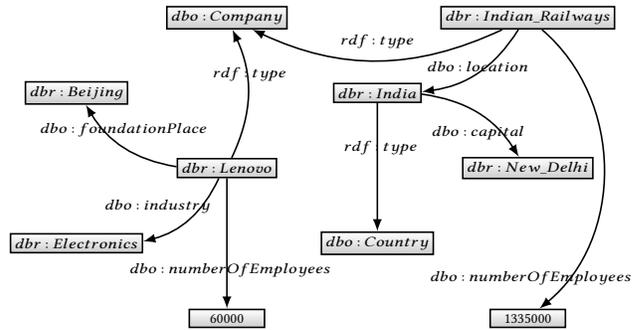
\begin{figure}[t]
\centering
\begin{tiny}
\begin {tikzpicture}
[-latex ,auto ,node distance =1 cm and 2cm ,
on grid ,
semithick ,
state/.style ={ top color =white , bottom color = black!20 ,
draw,black , text=black , minimum width =1 cm}]

\node[state] (B)  {$dbr:Lenovo$};

\node[state] (C) [below left =of B] {$dbr:Electronics$};

\node[state] (A) [above left=of B] {$dbr:Beijing$};

\node[state] (D) [above right=of A] {$dbo:Company$};

\node[state] (E) [above right=of B] {$dbr:India$};

\node[state] (F) [above right=of E] {$dbr:Indian\_Railways$};

\node[state] (G) [below right=of E] {$dbr:New\_Delhi$};

\node[state] (H) [below left=of G] {$dbo:Country$};

\node[state] (I) [below right =of C] {$60000$};

\node[state] (J) [below right=of H] {$1335000$};

\path (F) edge [bend left =50] node[above =-1.6 cm]
{$dbo:numberOfEmployees$\ \ \ \ \ \ \ \ \ \ \ \ \ \ \ \ \ \ \ \ \ \ \ \ \ \ \ \ \ \ \ \ \ \ \ \ \ \ \ \ \ \ \ \ \ } (J);

\path (B) edge [bend left =25] node[above =0.07 cm] 
{$dbo:foundationPlace$} (A);

\path (B) edge [bend left =25] node[above =0.01 cm]  
{$dbo:industry$\ \ \ \ \ \ \ \ \ \ \ \ \ \ \ \ \ \ \ \ } (C);

\path (B) 
edge [bend right =25] node[above =0.1 cm] 
{\ $rdf:type$} (D);

\path (F) 
edge [bend left =25] node[above =0.1 cm] 
{$rdf:type$} (D);

\path (F) 
edge [bend left =25] node[above =-0.05 cm] 
{$dbo:location$} (E);

\path (E) 
edge [bend left =25] node[below =0.1 cm] 
{$dbo:capital$} (G);

\path (E) 
edge  node[above =0.1 cm] 
{\ \ $rdf:type$} (H);

\path (B) 
edge  node[below =0.2 cm] 
{\ \ $dbo:numberOfEmployees$} (I);

\end{tikzpicture}
\end{tiny}
\caption{A Subgraph from the DBpedia.
} 
\label{fig:DBpediaSample}
\vspace{-2mm}
\end{figure}

{\textbf{Knowledge Graph (\emph{KG}):}} 
A knowledge graph is a directed graph $KG = \{V, E\}$ consisting of a set of vertices $V$ that represent entities, types, and literals, and a set of labeled edges $E$ that connect these vertices. 
RDF~\cite{RDF} is a popular representation model for KGs, which organizes data as a set of triples in the form \emph{$\langle s, p, o \rangle$} where \emph{$s$} refers to \emph{subject}, \emph{$p$} refers to \emph{predicate}, and \emph{$o$} refers to \emph{object}, such that $s, o \in V$, and $p \in E$. SPARQL~\cite{SPARQL} is the structured query language for querying RDF. \textcolor{black}{Figure \ref{fig:DBpediaSample} visualizes a sample subgraph from DBpedia~\cite{Auer2007}}.

\noindent{\textbf{Graph  Pattern (\emph{gp}):}} 
A Graph Pattern is a set of triple patterns. A triple pattern can be generated from a triple by replacing the $s$, $p$, or $o$ by a variable from the universal set of variables $U$. A query can include multiple graph patterns.

\noindent{\textbf{Query (\emph{q}):}}
A query is represented as the pair ($GP$, $SM$), where $GP$ is the set of graph patterns in the query and $SM$ is the set of solution modifiers. The query returns an answer set $A$ that (1) matches the given $GP$ from a knowledge graph $KG$, and (2) \textcolor{black}{modified to conform to the the solution modifiers $SM$ {(e.g., select, ordered by, distinct, limit and offset, etc.)}}. In SPARQL, a query with type $Select$ is used to select all, or a subset of, the variables bound in the set of subgraphs that matches $GP$. If type is $Ask$ instead of $Select$, it returns a Boolean answer; $true$ if $GP$ can be matched ($A$ is a non-empty set). 

\begin{sloppypar}
\noindent\textbf{Natural Language Question (\emph{nlq}):} A natural language question is represented as the \textcolor{black}{list} of all tokens of the question excluding white spaces and punctuation marks. Formally, \textcolor{black}{$nlq = [token_{1}, token_{2}, \hdots, token_{m}]$}, where $m$ is the number of tokens in the question. 
\end{sloppypar}

\noindent\textbf{Benchmark (\emph{B}):}
A benchmark $B = \{NLQ, Q , G\}$ consists of a set of natural language questions $NLQ$, an optional set of formal queries $Q$, and a set of gold standard answers $G$ such that $g_{i}$ is the set of gold standard answers for the question $nlq_{i}$, whose corresponding query is $q_{i}$.

\begin{exmp}
Following is an example of a natural language question that can be found in a benchmark ($nlq_{i}$): \textcolor{black}{$[$Which, companies, have, more, than, 1, million, employees, or, founded, in, Beijing$]$}. The query\footnote{The prefix dbr is bound to http://dbpedia.org/resource/\\
The prefix dbo is bound to http://dbpedia.org/ontology/\\
The prefix rdf is bound to http://www.w3.org/2000/01/rdf-schema\#} ($q_{i}$) that is associated with this question is:
\small
\begin{verbatim}
SELECT DISTINCT ?uri WHERE {
    ?uri a dbo:Company {
        ?uri dbo:numberOfEmployees ?n .
        FILTER ( ?n > 1000000 )
    } UNION { 
        ?uri dbo:foundationPlace dbr:Beijing. 
    }
} 
\end{verbatim}
\normalsize
The answers ($g_{i}$) to this query based on the subgraph in Figure~\ref{fig:DBpediaSample} are $\{$``Lenovo'', ``Indian Railways''$\}$.
\end{exmp}




\section{Overview of \system}
\label{sec:architecture}
 

 \begin{figure*}[t]
  \includegraphics[width=0.87\linewidth]{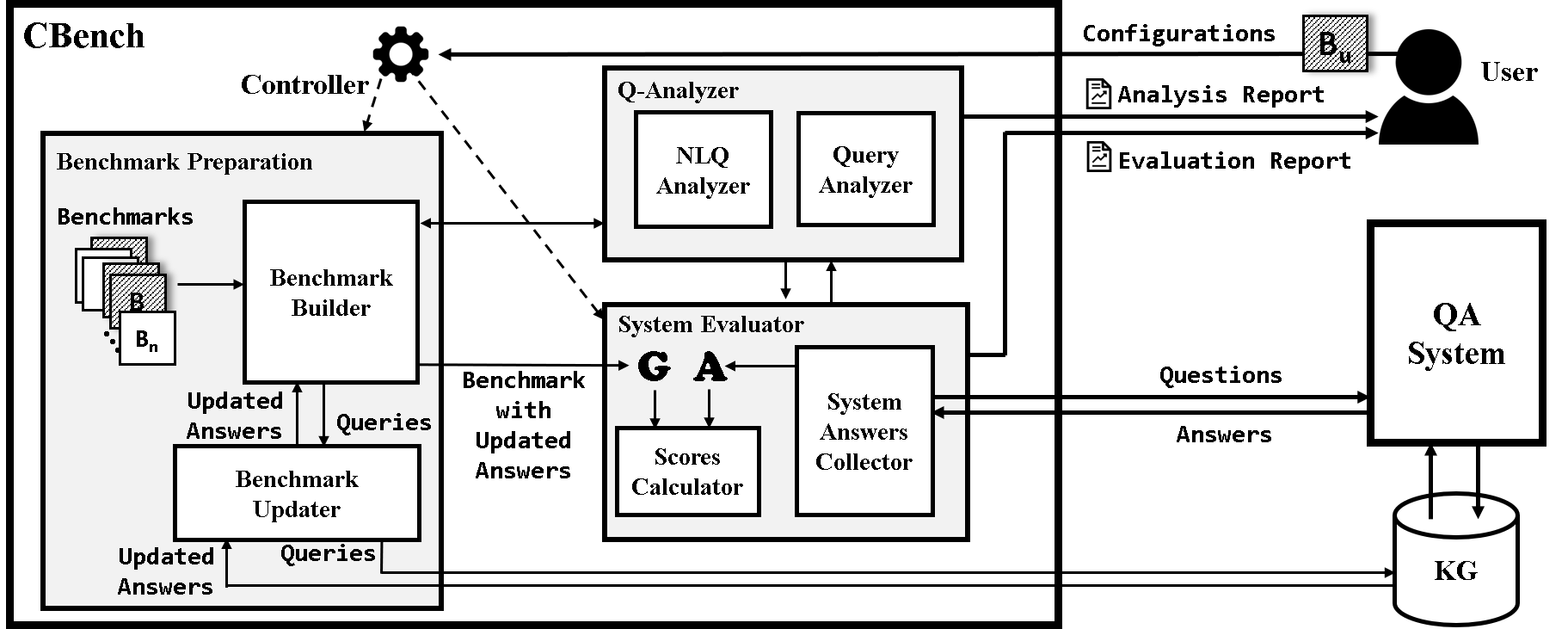}
  \caption{CBench Architecture.}
  \label{fig:cbenck}
\end{figure*}

\textcolor{black}{Figure~\ref{fig:cbenck} shows the architecture of \system, which can be used in two modes: (1) The Benchmark Analysis Mode, where \system can be used to perform a fine-grained analysis on the structured queries and the natural language questions on a set of benchmarks selected by the user, and (2) the QA Evaluation Mode, where \system can be used to evaluate a QA system over the user-selected benchmarks providing deeper insights on how the QA system is performing.\\
\textbf{Benchmark Analysis Mode}: \system includes 17 benchmarks from which the user can choose a subset for analysis. The user can also upload their own benchmarks to be included in the analysis. The Benchmark Builder passes the selected benchmarks (and the uploaded ones, if any) to the Q-Analyzer which carries out the syntactical and structural analysis of the queries (Section~\ref{sec:structuredAnalysis}), and the linguistic analysis (Section~\ref{sec:nl}). Finally, the Q-Analyzer returns the analysis report to the user.\\
\textbf{QA Evaluation Mode}: Just like the previous mode, the user selects a set of benchmarks and/or upload their own to evaluate the QA system. In addition, the user provides \system with a URL for an endpoint that \system can query. Other configuration parameters that are used in the evaluation (e.g., thresholds for calculating quality scores) are also chosen by the user prior to evaluation. To avoid the scenario where the selected benchmarks target different versions of the same KG (discussed in Section~\ref{sec:benchmarks}), the Benchmark Builder updates the answers of the queries in the selected benchmarks through the Benchmark Updater module, which queries the used KG that will be used for the evaluation to retrieve the updated answers. The updated benchmarks are then passed to the System Evaluator. The System Evaluator carries out three tasks: (1) Communicating with the QA system to collect the answers to the questions from the selected benchmarks, (2) calculating the micro, macro, and global F-1 scores (discussed in Section~\ref{sec:eval}), and (3) retrieving the fine-grained analysis of the processed questions from the Q-Analyzer. The System Evaluator then outputs an interactive report that includes the scores and the analysis of the processed questions to the user. The user can choose to focus on specific questions to view all their fine-grained properties. \system also finds other questions that are linguistically \emph{similar} to the selected questions (discussed in Section~\ref{sec:nl}). Using this feature, the user is able to quickly identify either common features or inconsistencies of unanswered or incorrectly answered questions, which will help the user to understand which components of their QA system to improve. The user can also use \system in the Debugging Mode, in which they can group questions/queries based on specific properties to evaluate their QA system in specific scenarios. For example, the user can choose to evaluate their QA system only on aggregate questions (e.g., How many) whose queries have a star-shape to investigate how their system deals with such questions.}

\textcolor{black}{The details of the configurations of \system, how to add a new benchmark, and the API's used for communication with the QA systems can be found in the system's repository\footnote{\textcolor{black}{\url{https://github.com/aorogat/CBench}}}.}

\section{Benchmarks in \system}
\label{sec:benchmarks}

\begin{table}[t]
    \captionof{table}{The benchmarks used in \system. DB refers to DBpedia, FB refers to Freebase, MB refers to MusicBrainz, WK refers to Wikidata, and LS refers to LinkedSpending. Benchmarks annotated with $\star$ do not include queries, and benchmarks annotated with $\dagger$ target only single-factoid questions.
}
\label{table:datasets}      
   \begin{tabular}{lclc}
\toprule
\multicolumn{1}{c}{\textbf{Benchmarks}} & 
\textbf{\#Qs} & 
\multicolumn{1}{c}{\textbf{KG}} & 
\multicolumn{1}{c}{\textbf{Version}} \\
\midrule
\emph{QALD-1} \cite{Unger2011}	     & 	199		 & 	DB, MB 	& 3.6 \\
\emph{QALD-2}	\cite{cabrio2012qakis}     & 	344	 	 & 	DB, MB	& 3.7 \\
\emph{QALD-3}	\cite{Cabrio2013}     & 	397	 	 & 	DB, MB	& 3.8 \\
\emph{QALD-4} \cite{Unger2014}	     & 	321	 	 & 	DB	& 3.9 \\
\emph{QALD-5} \cite{Unger2015}	     & 	334	 	 & 	DB	& 2014 \\
\emph{QALD-6}	\cite{Unger2016}     & 	431	 	 & 	DB, LS	& 10-2015\\
\emph{QALD-7}	\cite{Usbeck2017}     & 	530	 	 & 	DB, WD	& 04-2016\\
\emph{QALD-8}	\cite{Usbeck2018}     & 	315	 	 & 	DB, WD	& 10-2016\\
\emph{QALD-9}	\cite{ngomo20189th}     & 	408	 	 & 	DB	& 10-2016\\
\emph{LC-QuAD} \cite{trivedi2017lc}	 & 	4,998	 & 	DB	& 04-2016 
\\
\emph{WebQuestions} \cite{Berant2013} & 	5,810 & 	FB	& 09-08-2015 
\\
\emph{GraphQuestions} \cite{Su2016}  & 	5,166  & 	FB	& 06-2013 \\

\emph{SimpleQuestions}$\star\dagger$  \cite{bordes2015large}  	 & 	108,442		 & 	FB & FB2M, FB5M  \\
\emph{SimpleDBpediaQA}$\star\dagger$  \cite{Azmy2018}  	 & 	43,086	 	 & 	DB
& 10-2016\\
\emph{TempQuestions}$\star$  \cite{jia2018tempquestions}  	 & 	1,271		 & 	FB  & 09-08-2015\\
\emph{ComplexQuestions}$\star$ \cite{Abujabal2017}	 & 	150		 & 	FB	& 09-08-2015\\
\emph{ComQA}$\star$ \cite{abujabal-etal-2019-comqa}  	         & 	11,214    	 & 	Wikipedia	& -\\
	\bottomrule
\end{tabular}
\vspace{-2mm}
\end{table}

\begin{table}[!htbp]
\captionof{table}{QALDs Overlaps. \textbf{\#Qs} is the total number of all questions. \textbf{\#SU} is the number of \uline{S}ame questions but associated with \uline{U}pdated queries. \textbf{\#SS} is the number of \uline{S}ame questions and associated with the \uline{S}ame queries. \textbf{\#New} is the number of \uline{New} questions and associated with \uline{New} queries.}   
\label{table:qaldsOverlap}
\begin{tabular}{lp{1.3cm}p{1.3cm}p{1.3cm}p{1.3cm}}
\hline
       & \textbf{\#Qs}
       & 
       \textbf{\#SU}           & 
       \textbf{\#SS}           & 
       \textbf{\#New} \\
\hline
\emph{QALD-1} & 199           & -                         & -                        & 199                 \\
\emph{QALD-2} & 344           & 150                        & 0                        & 194                 \\
\emph{QALD-3} & 397           & 236                       & 52                        &109                   \\
\emph{QALD-4} & 321           & 146                       & 6                        & 169                 \\
\emph{QALD-5} & 334           & 27                        & 193                      & 114                 \\
\emph{QALD-6} & 431           & 317                       & 0                        & 114                 \\
\emph{QALD-7} & 530           & 262                       & 57                       & 211                 \\
\emph{QALD-8} & 315           & 174                       & 100                      & 41                  \\
\emph{QALD-9} & 408           & 231                       & 171                      & 6\\    
\hline
\end{tabular}
\end{table}

Before discussing the analysis of queries and natural language questions, we give an overview of the benchmarks that are included in \system. \system can also be augmented with other benchmarks provided by the user. Table~\ref{table:datasets} shows the benchmarks along with the number of questions in each and the KGs they target. \emph{QALD}\footnote{\url{http://qald.aksw.org}} is an annual evaluation campaign for question answering that started in 2011. Therefore, it includes 9 benchmarks (\emph{QALD-1} to \emph{QALD-9}). 
\textcolor{black}{
There are overlaps between these versions of QALD. As each question in any of the QALD benchmarks is associated with a query, there are cases where the natural language question is in a more recent benchmark but associated with a different query. Table~\ref{table:qaldsOverlap} shows the number of repeated natural language questions for each benchmark and how many of those have repeated queries (i.e., no change at all in the benchmark entry). The second column shows the total number of questions in each benchmark. While the third and fourth columns show the number of questions borrowed from previous versions, the fifth column shows how many questions were inserted for the first time in the corresponding version of \textit{QALD}. For example, \textit{QALD-9} has in total 408 questions. Only 6 questions are new. The other 402 questions are borrowed from all older versions, where 231 of the 402 questions use the same natural language questions but are associated with updated queries. The remaining 171 questions use the same natural language questions and the same queries. It is interesting to see that some questions are used in some version, left in a newer version and then reused again in a more recent version.
}

\emph{LC-QuAD} is a semi-automated question generation dataset. SPARQL templates are automatically generated and are converted into natural question templates. These general templates are manually transformed into natural language questions.
\emph{GraphQuestions} is a set of questions that are generated in two steps. First, generating a set of graph-structured logical patterns from the KG, then transforming them into natural questions with the help of human annotators.
\emph{WebQuestions} is a set of questions obtained from non-experts. These questions are collected based on suggestions from Google Suggest API. The questions with answers from Freebase are taken and annotated by Amazon Mechanical Turk workers then converted to SPARQL queries by experts.
\emph{SimpleQuestions} is generated by shortlisting the set of facts from Freebase that can be converted to informative questions. Then, these elected facts were sampled and passed to annotators to manually generate natural language questions whose answers are the entities in these facts.
\emph{SimpleDBpediaQA} is a subset of the \emph{SimpleQuestions} benchmark dataset created by mapping entities and predicates from Freebase to DBpedia. 
\emph{TempQuestions} consists exclusively of temporal questions. These questions are extracted from \emph{Free917}~\cite{cai-yates-2013-large}, \emph{WebQuestions} and \emph{ComplexQuestions}.
\emph{ComplexQuestions} is multi-constraints question-answer pairs that have some questions from \emph{WebQuestions} (596 questions) and some manually labelled questions.
\emph{ComQA} is a large benchmark that includes real questions taken from the WikiAnswers platform and present various challenging aspects such as compositionality, temporal reasoning, and comparisons.

\captionsetup[table]{skip=10pt}
\section{Analysis of Structured Queries}
\label{sec:structuredAnalysis}

\graphicspath{ {./figures/} }

Most of the benchmarks discussed in Section~\ref{sec:benchmarks} include structured queries that can be used to obtain the answers to the natural language questions in the benchmark. Inspired by prior work on the analysis of query logs of endpoints~\cite{Bonifati2017}, we discuss how the Query Analyzer processes the SPARQL queries by focusing on both their syntactically and structural properties to present them in the final report shown to the user. \textcolor{black}{We use \system's Benchmark Analysis Mode to process \emph{all} queries from the 12 benchmarks that include SPARQL queries to highlight the high-degree variations among them, then we give our insights in light of these variations.} In Section~\ref{sec:eval}, we use \system's QA Evaluation mode to experimentally show the effects of these variations on the accurate assessment of QA systems.

\vspace{-1mm}
\subsection{Syntactical Analysis of Queries}
\textcolor{black}{In this section, we study the syntactical properties of the queries, which include the frequency of query keywords, the number of triple patterns in the query, and the frequency usage of operators.}

\begin{table}[t]
\captionof{table}{Percentage of keyword occurrences in queries for each benchmark.}   
\label{table:keywords}
   \begin{tabular}{lrrrr}
\toprule
\textbf{Element} & \textbf{QALD} & \textbf{LC-QuAD} & \textbf{Web} & \textbf{Graph}  \\
\midrule
\emph{Select} 	 & 	91.63\%	 & 	91.52\% & 	100.00\%	 & 	100.00\%		\\
\emph{Ask}  	 & 	8.37\%	 & 	8.48\%	 & 	0.00\%	 & 	0.00\%	\\ \midrule
\emph{Distinct}   	 & 	76.65\%	 & 	91.52\%	 & 	99.98\%	 & 	0.00\%	\\
\emph{Limit}   	 & 	6.51\%	& 	0.00\% & 	0.02\%	 & 	0.00\%	 	\\
\emph{Offset}   	 & 	3.93\%	 & 	0.00\% & 	0.00\%	 & 	0.00\%	 	\\
\emph{Order By}    	 & 	5.99\%	 & 	0.00\% & 	0.02\%	 & 	0.00\%		\\ \midrule

\emph{And}  	 & 	51.65\%	 & 	70.67\% & 	37.65\%	 & 	41.75\%		\\
\emph{Filter} 	 & 	10.33\%	 & 	0.00\% & 	99.62\%	 & 	100.00\%		\\
\emph{Union}  	 & 6.10\% & 	0.00\%	 & 	0.36\%	 & 	0.00\%		\\
\emph{\textcolor{black}{Optional}} 	 & 	5.37\%	 & 	0.00\%	 & 	0.00\%	 & 	0.00\%	\\
\emph{Not Exists}  	 & 	0.21\%	 & 	0.00\%	 & 	0.00\%	 & 	0.00\%	\\
\emph{Minus}  	 & 	0.21\%	 & 	0.00\%	 & 	0.00\%	 & 	0.00\%	\\ \midrule
\emph{Aggregators}    	 & 	5.27\%
& 	0.00\%
& 	0.00\%	 & 	
20.17\% 
	\\
\emph{Group By}  	 & 	5.27\%	 & 	0.00\%	 & 	0.00\%	 & 	13.74\%	\\
\emph{Having}   	 & 	1.34\%	 & 	0.00\%	 & 	0.00\%	 & 	0.00\%	\\ \bottomrule
\end{tabular}
\vspace{-3mm}
\end{table}

\subsubsection{Query Keywords}
\textcolor{black}{We count the frequency of query keywords in the benchmarks. The results are shown in Table \ref{table:keywords}. Due to the lack of space, we combine all the queries from \emph{QALD-1} to \emph{QALD-9} and report them under \emph{QALD}.  In the case of repetitive questions, we consider the most recent query. After deduplication, the \emph{QALD} discussed here include 959 questions. We find that even within the 9 \emph{QALD} benchmarks, variations still exist. We will highlight these variations later in this section.} The first block in Table~\ref{table:keywords} reports the query types. In general, the majority of queries use the \emph{Select} keyword across all benchmarks (at least 91\% of queries). \emph{Ask} queries whose answers are either true (if a solution matches the graph pattern in the query) or false (otherwise) represent 8.37\% and 8.48\% of the queries in \emph{QALD} and \emph{LC-QuAD}, respectively. \emph{GraphQuestions} and \emph{WebQuestions} do not include queries that use the keyword \emph{Ask}.

The second block in Table \ref{table:keywords} includes the keywords used as solution modifiers in the queries. \textcolor{black}{We ignored the Reduced modifier as it is never used in all benchmarks}. We notice that the majority of the queries use the \emph{Distinct} keyword in all benchmarks except \emph{Graph\-Questions}, which is the only benchmark that does not use any \textcolor{black}{of the four} solution modifiers. The \emph{Limit}, \emph{Offset} and \emph{Ordered-By} keywords are not frequently used in \emph{QALD}, and almost  non-existent in other benchmarks.

The third block in Table \ref{table:keywords} has keywords used to describe the graph patterns as described in Section~\ref{sec:preliminaries}. \textcolor{black}{The \emph{And} operator, which represents the conjunctions of triple patterns, is often used across all benchmarks.} However, there is a large variation in the percentage of queries that use the keyword, where the minimum percentage is 37.65\% (\emph{WebQuestions}), and the maximum percentage is 70.67\% (\emph{LC-QuAD}). The \emph{Filter} keyword demonstrates an interesting case, where it is used in almost all questions in \emph{WebQuestions} and \emph{GraphQuestions}, but with a much smaller percentage in \emph{QALD} and not used at all in \emph{LC-QuAD}. The remaining keywords (\emph{Union}, \emph{\textcolor{black}{Optional}}, \emph{Not Exists}, and \emph{Minus}) are used in a small percentage of queries across all benchmarks. But again, there is a high degree of variation across all benchmarks in their usage for these keywords.

Lastly, \emph{Aggregators} (e.g., \emph{count}, \emph{max}, etc.) and \emph{Group-By} keywords are used only in \emph{QALD} and \emph{GaphQuestions}. The keyword \emph{Having} is used in \emph{QALD} only. 

\begin{table*}[t]
\captionof{table}{Percentage of keyword occurrences in queries for each QALD benchmark.}   
\label{table:keywordsQALD}
\begin{tabular}{lccccccccc}
\toprule
\textbf{Element}     & \textbf{QALD-1}  & \textbf{QALD-2}  &\textbf{QALD-3}  & \textbf{QALD-4}  & \textbf{QALD-5}  & \textbf{QALD-6}  & \textbf{QALD-7}  & \textbf{QALD-8}  & \textbf{QALD-9}  \\
\midrule
\emph{Select}      & 90.48\% & 89.26\%      & 90.54\%     & 91.13\% & 91.78\% & 92.77\% & 88.04\% & 89.17\% & 90.61\% \\
\emph{Ask}         & 9.52\%  & 10.74\%      & 9.46\%      & 8.87\%  & 8.22\%  & 7.23\%  & 11.96\% & 10.83\% & 9.39\%  \\
\midrule
\emph{Distinct}    & 38.69\% & 80.00\%      & 86.25\%     & 91.13\% & 89.14\% & 86.51\% & 77.20\% & 67.83\% & 76.40\% \\
\emph{Limit}       & 4.17\%  & 4.44\%       & 4.30\%      & 8.06\%  & 9.21\%  & 8.19\%  & 7.45\%  & 11.15\% & 9.64\%  \\
\emph{Offset}      & 0.00\%  & 1.85\%       & 2.01\%      & 6.05\%  & 8.55\%  & 7.71\%  & 4.29\%  & 5.41\%  & 6.09\%  \\
\emph{Order By}    & 4.17\%  & 4.44\%       & 4.30\%      & 8.06\%  & 9.21\%  & 8.19\%  & 7.45\%  & 9.55\%  & 8.88\%  \\
\midrule
\emph{And}         & 79.76\% & 64.81\%      & 60.74\%     & 50.81\% & 50.33\% & 46.75\% & 39.28\% & 35.99\% & 42.89\% \\
\emph{Filter}      & 39.29\% & 54.44\%      & 9.17\%      & 8.06\%  & 6.58\%  & 4.34\%  & 9.71\%  & 3.18\%  & 4.82\%  \\
\emph{Union}       & 7.74\%  & 7.04\%       & 5.73\%      & 8.87\%  & 9.87\%  & 8.19\%  & 4.97\%  & 1.59\%  & 6.60\%  \\
\emph{Optional}    & 38.10\% & 50.00\%      & 0.86\%      & 0.40\%  & 0.33\%  & 0.24\%  & 6.09\%  & 0.32\%  & 0.25\%  \\
\emph{Not Exists}  & 0.00\%  & 0.00\%       & 0.00\%      & 0.00\%  & 0.00\%  & 0.48\%  & 0.45\%  & 0.64\%  & 0.00\%  \\
\emph{Minus}       & 0.00\%  & 0.00\%       & 0.00\%      & 0.81\%  & 0.00\%  & 0.00\%  & 0.00\%  & 0.00\%  & 0.00\%  \\
\midrule
\emph{Aggregators} & 2.98\%  & 3.33\%       & 4.30\%      & 1.61\%  & 2.63\%  & 6.02\%  & 5.64\%  & 5.10\%  & 4.82\%  \\
\emph{Group By}    & 2.98\%  & 3.33\%       & 4.30\%      & 1.61\%  & 2.63\%  & 6.02\%  & 5.64\%  & 5.10\%  & 4.82\%  \\
\emph{Having}      & 0.60\%  & 2.22\%       & 2.29\%      & 0.81\%  & 0.66\%  & 0.72\%  & 0.68\%  & 0.32\%  & 0.76\% \\
\bottomrule
\end{tabular}
\end{table*}
With respect to the 9 \emph{QALD} benchmarks, variations also exist although they come from the same organization as shown in Table~\ref{table:keywordsQALD}. For example, the 
\emph{Distinct} keyword occurs in only 38.69\% of the \emph{QALD-1} queries , whereas in the other \emph{QALD} benchmarks, the \emph{Distinct} keyword occurs in between 67\% to 91\% of the queries. 
The \emph{Limit} keyword occurrences steadily increase from 4.17\% in \emph{QALD-1} to 9.64\% in \emph{QALD-9}. The \emph{Offset} keyword is approximately never used in \emph{QALD-1} to \emph{QALD-3}, but occurs more frequently in the other \emph{QALD} benchmarks (from 4\% to 8\%). \textcolor{black}{The \emph{Filter} and \emph{Optional} keywords occur frequently in \emph{QALD-1} and \emph{QALD-2} (39.29\% and 54.44\% of the queries, respectively), while this percentage is significantly declined in other \emph{QALD} benchmarks (does not exceed 11\%).}

\subsubsection{Number of \textcolor{black}{Triple Patterns}}
 
 \begin{figure}[t]
  \includegraphics[width=0.95\linewidth]{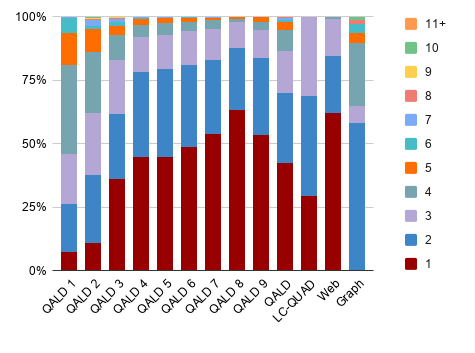}
  \vspace{-7mm}
  \caption{Percentage of queries exhibiting different number of \textcolor{black}{triple patterns} for each benchmark.
  }
  \label{fig:triples}
  \vspace{-3mm}
\end{figure}

The Query Analyzer also counts the number of \textcolor{black}{triple patterns} in all the graph patterns of the queries as a measure of the size of the queries in the benchmarks. The total number of \textcolor{black}{triple patterns} has been computed and categorized from 1 to 11+ \textcolor{black}{triple patterns} for every benchmark. Figure~\ref{fig:triples} shows that the queries with a low number of \textcolor{black}{triple patterns} (from 1 to 3) are dominant in all benchmarks except  the \emph{GraphQuestions} benchmark, in which there are no queries with 1 \textcolor{black}{triple pattern}. Noticeably, the earlier versions of \emph{QALD} and the \emph{GraphQuestions} benchmarks include longer queries when compared to the other benchmarks.

\subsubsection{Query Operators}

\begin{table}[t]
\captionof{table}{\textcolor{black}{The frequency of the operators used in queries: Filter (F), And (A), Optional (O), and Union (U).}}
        \label{table:operators}
\begin{tabular}{lrrrr}
\toprule
\textbf{Operators} & \textbf{QALD} & \textbf{LC-QuAD} & \textbf{Web} & \textbf{Graph}   \\ \midrule
$none$ 	 & 	42.25\%	 & 	29.33\%	 & 	0.09\%	 & 	0.00\%	\\
$F$  	 & 0.00\%  & 	0.00\%	 & 	62.19\%	 & 	58.25\%		\\
$A$     	 & 42.87\%	 & 	70.67\% & 	0.17\%	 & 	0.00\%		\\
$A, F$    	 & 4.65\%	 & 	0.00\% & 	37.19\%	 & 	41.75\%		\\ \midrule
\textbf{$CPF$}    	 & 	\textbf{89.77\%}	  & 	\textbf{100.00\%} & 	\textbf{99.64\%}	 & 	\textbf{100.00\%}		\\ \midrule
$O$      	 & 	0.00\%	 & 	0.00\%	 & 	0.00\%	 & 	0.00\%	\\
$O, F$    	 & 2.58\%	 & 	0.00\%	 & 	0.00\%	 & 	0.00\%	\\
$A, O$   	 & 0.10\%	 & 	0.00\%	 & 	0.00\%	 & 	0.00\%	\\
$A, O, F$ 	 & 1.45\%	 & 	0.00\%	 & 	0.00\%	 & 	0.00\%	\\ \midrule
\textbf{$CPF + O$} 	 & 	\textbf{+4.13\%}	 & 	\textbf{+0.00\%}	 & 	\textbf{+0.00\%}	 & 	\textbf{+0.00\%}	\\ \midrule
$U$      	 & 2.48\%  & 	0.00\%	 & 	0.07\%	 & 	0.00\%		\\
$U, F$     	 & 0.10\%	 & 	0.00\%	 & 	0.00\%	 & 	0.00\%	\\
$A, U$     	 & 1.96\%  & 	0.00\%	 & 	0.05\%	 & 	0.00\%		\\
$A, U, F$    	 & 	0.31\% & 	0.00\%	 & 	0.24\%	 & 	0.00\%		\\ \midrule
\textbf{$CPF + U$} 	 & 	\textbf{+4.86\%}  & 	\textbf{+0.00\%}	 & 	\textbf{+0.36\%}	 & 	\textbf{+0.00\%}		\\ \bottomrule
\end{tabular}
\end{table}

We also study the co-occurrences of the \emph{Filter}, \emph{And}, \emph{Union}, and \emph{\textcolor{black}{Optional}} operators in the queries. The results can be found in Table \ref{table:operators}. The first block of the table shows the queries with graph patterns that have \textcolor{black}{triple patterns} with no operators \textcolor{black}{(with a single triple pattern without filter)}, with \emph{Filter} only, with \emph{And} only, or with both \emph{Filter} and \emph{And} operators. The next row has their subtotal \textcolor{black}{(\emph{\ul{c}onjunctive \ul{p}atterns with \ul{f}ilters}  or \emph{CPF})}. Most of the queries (from 89.77\% to 100.00\%) in all benchmarks are \emph{CPF} queries. However, the distribution of queries using the combinations of the two operators varies. 

The second block in the table shows the co-occurrences of the \emph{\textcolor{black}{Optional}} operator with different types of \emph{CPF} queries. This operator is only used in the \emph{QALD} benchmarks with an increase of +4.13\% in the relative size. 

\begin{table*}[]
\captionof{table}{\textcolor{black}{The frequency of the operators used in QALDs queries: Filter (F), And (A), Optional (O), and Union (U).}}
        \label{table:operators_Qalds}
\begin{tabular}{lrrrrrrrrr}
\toprule
        \multicolumn{1}{l}{\textbf{Operators}}
        & \multicolumn{1}{l}{\textbf{QALD-1}} & \multicolumn{1}{l}{\textbf{QALD-2}} & \multicolumn{1}{l}{\textbf{QALD-3}} & \multicolumn{1}{l}{\textbf{QALD-4}} & \multicolumn{1}{l}{\textbf{QALD-5}} & \multicolumn{1}{l}{\textbf{QALD-6}} & \multicolumn{1}{l}{\textbf{QALD-7}} & \multicolumn{1}{l}{\textbf{QALD-8}} & \multicolumn{1}{l}{\textbf{QALD-9}} \\
\midrule
\textbf{$None$}    & 7.14\%                              & 10.37\%                             & 35.53\%                             & 44.76\%                             & 44.41\%                             & 48.43\%                             & 53.50\%                             & 63.06\%                             & 53.05\%                             \\
\textbf{$F$}       & 0.00\%                              & 0.37\%                              & 0.00\%                              & 0.00\%                              & 0.00\%                              & 0.00\%                              & 0.00\%                              & 0.00\%                              & 0.00\%                              \\
\textbf{$A$}       & 50.60\%                             & 35.19\%                             & 50.14\%                             & 39.92\%                             & 40.46\%                             & 39.76\%                             & 32.96\%                             & 32.17\%                             & 36.04\%                             \\
\textbf{$A, F$}    & 3.57\%                              & 3.70\%                              & 7.74\%                              & 6.05\%                              & 4.93\%                              & 3.37\%                              & 3.61\%                              & 2.87\%                              & 4.06\%                              \\
\midrule
\textbf{$CPF$}     & \textbf{61.31\%}                             & \textbf{49.63\%}                             & \textbf{93.41\%}                             & \textbf{90.73\%}                             & \textbf{89.80\%}                             & \textbf{91.57\%}                             & \textbf{90.07\%}                             & \textbf{98.09\%}                             & \textbf{93.15\%}                             \\
\midrule
\textbf{$O$}       & 0.60\%                              & 0.00\%                              & 0.00\%                              & 0.00\%                              & 0.00\%                              & 0.00\%                              & 0.00\%                              & 0.00\%                              & 0.00\%                              \\
\textbf{$O, F$}    & 7.74\%                              & 20.37\%                             & 0.57\%                              & 0.40\%                              & 0.33\%                              & 0.24\%                              & 4.51\%                              & 0.32\%                              & 0.25\%                              \\
\textbf{$A, O$}    & 1.79\%                              & 0.00\%                              & 0.00\%                              & 0.00\%                              & 0.00\%                              & 0.00\%                              & 0.00\%                              & 0.00\%                              & 0.00\%                              \\
\textbf{$A, O, F$} & 20.83\%                             & 22.96\%                             & 0.29\%                              & 0.00\%                              & 0.00\%                              & 0.00\%                              & 0.45\%                              & 0.00\%                              & 0.00\%                              \\
\midrule
\textbf{$CPF + O$} & \textbf{+30.95\%}                             & \textbf{+43.33\%}                             & \textbf{+0.86\%}                              & \textbf{+0.40\%}                              & \textbf{+0.33\%}                              & \textbf{+0.24\%}                              & \textbf{+4.97\%}                              & \textbf{+0.32\%}                              & \textbf{+0.25\%}                              \\
\midrule
\textbf{$U$}       & 0.60\%                              & 0.00\%                              & 3.15\%                              & 3.63\%                              & 4.28\%                              & 4.34\%                              & 1.58\%                              & 0.64\%                              & 3.55\%                              \\
\textbf{$U, F$}      & 0.00\%                              & 0.00\%                              & 0.00\%                              & 0.40\%                              & 0.66\%                              & 0.24\%                              & 0.00\%                              & 0.00\%                              & 0.25\%                              \\
\textbf{$A, U$}      & 0.00\%                              & 0.00\%                              & 2.01\%                              & 3.63\%                              & 4.28\%                              & 3.13\%                              & 2.26\%                              & 0.96\%                              & 2.54\%                              \\
\textbf{$A, U, F$}     & 0.00\%                              & 0.37\%                              & 0.57\%                              & 1.21\%                              & 0.66\%                              & 0.48\%                              & 0.00\%                              & 0.00\%                              & 0.25\%                              \\
\midrule
\textbf{$CPF + U$} & \textbf{+0.60\%}                             & \textbf{+0.37\%}                              & \textbf{+5.73\%}                              & \textbf{+8.87\%}                              & \textbf{+9.87\%}                              & \textbf{+8.19\%}                              & \textbf{+3.84\%}                             & \textbf{+1.59\%}                              & \textbf{+6.60\%}     \\                        
\bottomrule
\end{tabular}
\end{table*}
Finally, The third block of the table shows that the \emph{Union} operator is not used in \emph{GraphQuestions} and \emph{LC-QuAD}, and is rarely used in \emph{WebQuestions}, while it is more frequently used in \emph{QALD}. \textcolor{black}{For the QALD benchmarks, there are other combinations that are rare and therefore not included in this table. For example, (O,U), (O,F,U), (A,O,U), etc.}. Table \ref{table:operators_Qalds} shows the results of each QALD version separately.

\subsection{Structural Analysis}
In addition to the syntactical analysis of the queries in the benchmarks, we also analyze the structural shapes of the queries of the following types of queries: 1.~Conjunctive queries that can use only the \emph{And} operator, denoted by \emph{CQ}. 2.~Conjunctive queries that can use both \emph{And} and \emph{Filter} operators, denoted by \emph{CQ\textsubscript{F}}. 3.~Conjunctive queries that can use \emph{And}, \emph{Filter} and \emph{\textcolor{black}{Optional}} operators, denoted by \emph{CQ\textsubscript{OF}}.

 \begin{figure}[t]
  \includegraphics[width=0.9\linewidth]{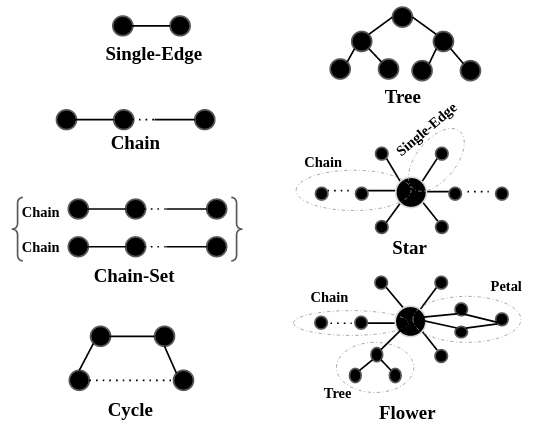}
  \caption{The different shapes recognized by \system}
  \label{fig:shapes}
  \vspace{-3mm}
\end{figure}

\begin{table*}[t]
\captionof{table}{Cumulative shape analysis of \textit{CQ}, \textit{CQ\textsubscript{F}}, \textit{CQ\textsubscript{OF}} across all benchmarks.}
\label{table:shapes}
\begin{tabular}{lrrrrrrrrrrrr}
\toprule
 & \multicolumn{3}{c}{\textbf{QALD}} 
 & \multicolumn{3}{c}{\textbf{LC-QuAD}}  
 & \multicolumn{3}{c}{\textbf{Web}} 
 & \multicolumn{3}{c}{\textbf{Graph}} 

 \\ 

 \cmidrule(lr){2-4}\cmidrule(lr){5-7}
 \cmidrule(lr){8-10}\cmidrule(lr){11-13}

 \textbf{Shape}
 &\multicolumn{1}{c}{\textit{CQ}}        
 &\multicolumn{1}{c}{\textit{CQ\textsubscript{F}}}
 &\multicolumn{1}{c}{\textit{CQ\textsubscript{OF}}}           
 &\multicolumn{1}{c}{\textit{CQ}}        
 &\multicolumn{1}{c}{\textit{CQ\textsubscript{F}}}
 &\multicolumn{1}{c}{\textit{CQ\textsubscript{OF}}}  
 
 &\multicolumn{1}{c}{\textit{CQ}}        
 &\multicolumn{1}{c}{\textit{CQ\textsubscript{F}}}
 &\multicolumn{1}{c}{\textit{CQ\textsubscript{OF}}}  
 
 &\multicolumn{1}{c}{\textit{CQ}}        
 &\multicolumn{1}{c}{\textit{CQ\textsubscript{F}}}
 &\multicolumn{1}{c}{\textit{CQ\textsubscript{OF}}}

 \\ \hline
Single-Edge 	&	 
50.82\%    	&	 46.88\%    	&	 45.13\% 	&
29.35\%    	&	 29.35\%    	&	 29.35\%  	&
40.00\%    	&	 63.29\%    	&	 63.29\% 	&	 
0\%    	&	 0.00\%    	&	 0.00\% 		 

\\
Chain       	&	 
83.77\%    	&	 80.79\%    	&	 80.20\% 	&
72.93\%    	&	 72.91\%    	&	 72.91\%  	&
80.00\%    	&	 90.98\%    	&	 90.98\% 	&	 
0\%    	&	 59.56\%    	&	 59.56\% 		 

\\
Chain-Set   	&	 
83.90\%    	&	 81.60\%    	&	 80.97\% 	&
72.93\%    	&	 72.91\%    	&	 72.91\%  	&
80.00\%    	&	 90.98\%    	&	 90.98\% 	&	 
0\%    	&	 59.56\%    	&	 59.56\% 		 

\\
Star        	&	 
14.84\%    	&	 15.86\%    	&	 16.59\% 	&
27.05\%    	&	 27.06\%    	&	 27.06\%  	&
20.00\%    	&	 8.61\%    	&	 8.61\% 	&	 
0\%    	&	 30.70\%    	&	 30.70\% 	

\\
Tree        	&	 
99.75\%    	&	 97.80\%    	&	 97.90\% 	&
100.00\%    	&	 100.00\%    	&	 100.00\%  	&
100.00\%    	&	 99.66\%    	&	 99.66\% 	&	
0\%    	&	 97.74\%    	&	 97.74\% 		 

\\
Forest      	&	 
100.00\%    	&	 99.54\%    	&	 99.56\%  	&	 
100.00\%    	&	 100.00\%    	&	 100.00\%  	&
100.00\%    	&	 99.93\%    	&	 99.93\%  	&	
0\%    	&	 97.74\%    	&	 97.74\%  		 

\\ 
Cycle       	&	 
0.13\%    	&	 0.12\%    	&	 0.11\% 	&	 
0.00\%    	&	 0.00\%    	&	 0.00\%  	&
0.00\%    	&	 0.32\%    	&	 0.32\% 	&	 
0\%    	&	 0.00\%    	&	 0.00\% 		 

\\
Flower      	&	 
99.87\%    	&	 98.15\%    	&	 98.23\% 	&
100.00\%    	&	 100.00\%    	&	 100.00\% &
100.00\%    	&	 99.98\%    	&	 99.98\% 	&	
0\%    	&	 99.33\%    	&	 99.33\% 		 

\\ \bottomrule


\end{tabular}
\vspace{-2mm}
\end{table*}
\normalsize

\textcolor{black}{\system identifies eight different shapes of queries. Figure~\ref{fig:shapes} illustrate these shapes\footnote{\textcolor{black}{Examples of questions, their corresponding queries, and their structural shapes can be viewed in the repository.}}.} The \emph{Single-Edge} shape has only one edge. The \emph{Chain} shape with length $n$ is a series of edges \{$x_0$,$x_1$\}, \{$x_1$,$x_2$\}, $\hdots$ \{$x_{n-1}$,$x_n$\}. The \emph{Cycle} shape is like the \emph{Chain} shape except that the first node in the chain is the same as the last node. The \emph{Chain-Set} shape is a set of one or more unconnected chains. The \emph{Tree} shape can have any connected nodes keeping only one path between any two nodes. The \emph{Star} shape is a special case of the \emph{Tree} shape where there exists exactly one node with more than 2 neighbors. The \emph{Flower} shape is the graph that has a node that is connected to at least one attachment that could have any of the following three shapes: \emph{Chain}, \emph{Tree} and \emph{Petal}, where the \emph{Petal} is a graph with two or more disjoint paths between a source node and a destination node. The \emph{Forest} shape includes a set of unconnected \emph{trees}.
It is worth noting that some shapes subsume other shapes. For example, the \emph{Tree} shape subsumes the \emph{Chain} shape, which subsumes the \emph{Single-Edge} shape. 

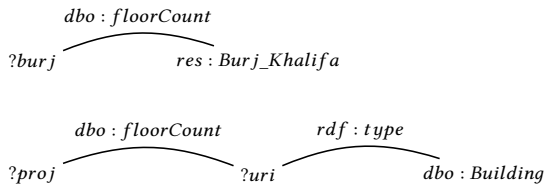
\begin{figure}[t]
\centering
\begin{footnotesize}
\begin {tikzpicture}
[-latex ,auto ,node distance =1.5cm and 3cm ,
on grid , 
semithick ]

\node (A)  {$?burj$};
\node (B) [right =of A] {$res:Burj\_Khalifa$};

\node (C) [below =of B] {$?uri$};
\node (D) [right =of C] {$dbo:Building$};
\node (E) [left =of C] {$?proj$};

\path (A) edge [-, bend left =20] node[above =0.0 cm] 
{$dbo:floorCount$} (B);
\path (C) edge [-, bend left =20] node[above =0.0 cm] 
{$rdf:type$} (D);
\path (C) edge [-, bend right =20] node[above =0.0 cm] 
{$dbo:floorCount$} (E);

\end{tikzpicture}
\end{footnotesize}
\caption{A Chain-Set example from QALD-9.
} 
\label{fig:chainset}
\end{figure}

\textcolor{black}{The set-shapes questions (e.g., \emph{Chain-Set} and \emph{Forest}) are used in NLQ where a comparison between 2 variables is required.  Figure \ref{fig:chainset} is an example from QALD-9 which visualizes the \textit{Chain-Set} query:}
\small
\begin{verbatim}
SELECT DISTINCT ?uri WHERE { 
    res:Burj_Khalifa dbo:floorCount ?burj . 
    ?uri rdf:type dbo:Building . 
    ?uri dbo:floorCount ?proj 
    FILTER ( ?proj < ?burj ) 
} ORDER BY DESC(?proj) LIMIT 1
\end{verbatim}
\normalsize
\textcolor{black}{This query corresponds to the NLQ: \textit{Which building after the Burj Khalifa has the most floors?} As shown in the query, there are two chains. The first chain (a single triple pattern) finds the number of floors of Burj Khalifa. The second chain (two triple patterns) finds all buildings and their number of floors.}

Table~\ref{table:shapes} shows the distribution of query shapes across all benchmarks. Again, due to the lack of space, we combine all 9 benchmarks of \emph{QALD} into one column. The table shows that queries with the \emph{Single-Edge} shape represent at least 45.13\% of the queries in almost all benchmarks. This shape corresponds to simple-factoid questions. The only exception is the \emph{GraphQuestions} benchmark, which does not include any \emph{Single-Edge} queries.

It is interesting to see that a high percentage of the queries have the \emph{Chain} shape. With the exception of \emph{GraphQuestions}, this shape includes at least 72\% of queries in all benchmarks. This observation along with the information from Figure~\ref{fig:triples} indicate that a high percentage of the queries have a \emph{Chain} shape with a relatively short length of the chain. In contrast, the \emph{Chain-Set} shape distribution has almost the same percentage as that of the \emph{Chain} shape, which indicates that queries that have the \emph{Chain-Set} shape, but not the \emph{Chain} shape are rare in all benchmarks. The same observation applies to the \emph{Tree} and \emph{Forest} shapes, indicating rare occurrences of \emph{Forest} shapes that are not trees. 
The \emph{Star} shape distribution ranges between 8.61\% and 30.7\% across all benchmark. The \emph{Cycle} shape is poorly represented in all benchmarks as every benchmark has from one to two queries or no queries for this shape.

\textcolor{black}{ For the different versions of QALD, 
Tables~\ref{tab:shapes_cq}, \ref{tab:shapes_cqf}, and \ref{tab:shapes_cqof} show the distribution of the discussed shapes for $CQ$, $CQ_F$, and $CQ_{OF}$ queries respectively.}

\begin{table*}[]
\caption{Cumulative shape analysis of $CQ$ across all versions of QALD benchmarks.}
\label{tab:shapes_cq}
\begin{tabular}{llllllllll}
\toprule
Shape & \textbf{QALD-1} & \textbf{QALD-2} & \textbf{QALD-3} & \textbf{QALD-4} & \textbf{QALD-5} & \textbf{QALD-6} & \textbf{QALD-7} & \textbf{QALD-8} & \textbf{QALD-9} \\
\midrule
\emph{Single-Edge} & 13.64\% & 24.56\% & 42.91\% & 53.92\% & 53.15\% & 55.71\% & 63.03\% & 66.78\% & 60.53\% \\
\emph{Chain}       & 50\%    & 54.39\% & 78.89\% & 92.16\% & 90.94\% & 91.64\% & 89.89\% & 91.1\%  & 91.81\% \\
\emph{Chain-Set}   & 50\%    & 54.39\% & 78.89\% & 92.16\% & 90.94\% & 91.64\% & 90.16\% & 91.1\%  & 91.81\% \\
\emph{Star}        & 47.73\% & 42.11\% & 19.03\% & 7.35\%  & 8.27\%  & 7.8\%   & 9.31\%  & 8.56\%  & 7.89\%  \\
\emph{Tree}        & 98.86\% & 100\%   & 99.65\% & 99.51\% & 99.21\% & 99.72\% & 99.73\% & 100\%   & 99.71\% \\
\emph{Forest}      & 98.86\% & 100\%   & 100\%   & 100\%   & 100\%   & 100\%   & 100\%   & 100\%   & 100\%   \\
\emph{Cycle}       & 0\%     & 0\%     & 0.35\%  & 0.49\%  & 0.39\%  & 0.28\%  & 0\%     & 0\%     & 0.29\%  \\
\emph{Flower}      & 100\%   & 100\%   & 100\%   & 100\%   & 99.61\% & 100\%   & 99.73\% & 100\%   & 100\%  \\
\bottomrule
\end{tabular}
\end{table*}

\begin{table*}[]
\caption{Cumulative shape analysis of $CQ_F$ across all versions of QALD benchmarks.}
\label{tab:shapes_cqf}
\begin{tabular}{llllllllll}
\toprule
Shape & \textbf{QALD-1} & \textbf{QALD-2} & \textbf{QALD-3} & \textbf{QALD-4} & \textbf{QALD-5} & \textbf{QALD-6} & \textbf{QALD-7} & \textbf{QALD-8} & \textbf{QALD-9} \\
\midrule
\emph{Single-Edge} & 11.65\% & 21.64\% & 38.04\% & 49.11\% & 49.63\% & 53.17\% & 59.7\%  & 63.93\% & 57.02\% \\
\emph{Chain}       & 49.51\% & 51.49\% & 75.77\% & 88.84\% & 87.5\%  & 89.15\% & 86.65\% & 89.18\% & 89.53\% \\
\emph{Chain-Set}   & 50.49\% & 51.49\% & 76.99\% & 89.73\% & 88.24\% & 89.68\% & 87.91\% & 89.51\% & 89.81\% \\
\emph{Star}        & 47.57\% & 40.3\%  & 19.33\% & 8.48\%  & 9.93\%  & 8.99\%  & 10.33\% & 9.18\%  & 8.82\%  \\
\emph{Tree}        & 98.06\% & 95.52\% & 96.63\% & 97.32\% & 97.43\% & 98.41\% & 97.48\% & 98.69\% & 98.35\% \\
\emph{Forest}      & 99.03\% & 97.76\% & 99.08\% & 100\%   & 100\%   & 100\%   & 100\%   & 100\%   & 100\%   \\
\emph{Cycle}       & 0\%     & 0\%     & 0.31\%  & 0.45\%  & 0.37\%  & 0.26\%  & 0\%     & 0\%     & 0.28\%  \\
\emph{Flower}      & 99.03\% & 96.27\% & 97.55\% & 97.77\% & 97.79\% & 98.68\% & 97.48\% & 98.69\% & 98.62\% \\
\bottomrule
\end{tabular}
\end{table*}

\begin{table*}[]
\caption{Cumulative shape analysis of $CQ_{OF}$ across all versions of QALD benchmarks.}
\label{tab:shapes_cqof}
\begin{tabular}{llllllllll}
\toprule
Shape & \textbf{QALD-1} & \textbf{QALD-2} & \textbf{QALD-3} & \textbf{QALD-4} & \textbf{QALD-5} & \textbf{QALD-6} & \textbf{QALD-7} & \textbf{QALD-8} & \textbf{QALD-9} \\
\midrule
\emph{Single-Edge} & 7.74\%  & 11.55\% & 38.3\%  & 49.33\% & 49.82\% & 53.3\%  & 57.04\% & 64.05\% & 57.14\% \\
\emph{Chain}       & 42.58\% & 51\%    & 75.68\% & 88.89\% & 87.55\% & 89.18\% & 86.87\% & 89.22\% & 89.56\% \\
\emph{Chain-Set}   & 43.87\% & 51\%    & 76.9\%  & 89.78\% & 88.28\% & 89.71\% & 88.07\% & 89.54\% & 89.84\% \\
\emph{Star}        & 52.26\% & 42.23\% & 19.45\% & 8.44\%  & 9.89\%  & 8.97\%  & 10.26\% & 9.15\%  & 8.79\%  \\
\emph{Tree}        & 96.13\% & 96.02\% & 96.66\% & 97.33\% & 97.44\% & 98.42\% & 97.61\% & 98.69\% & 98.35\% \\
\emph{Forest}      & 99.35\% & 98.8\%  & 99.09\% & 100\%   & 100\%   & 100\%   & 100\%   & 100\%   & 100\%   \\
\emph{Cycle}       & 0\%     & 0\%     & 0.3\%   & 0.44\%  & 0.37\%  & 0.26\%  & 0\%     & 0\%     & 0.27\%  \\
\emph{Flower}      & 97.42\% & 96.41\% & 97.57\% & 97.78\% & 97.8\%  & 98.68\% & 97.61\% & 98.69\% & 98.63\%\\
\bottomrule
\end{tabular}
\end{table*}

\vspace{-1mm}
\subsection{\textcolor{black}{Insights on Benchmark Selection (1)}}
\label{subsec:insights1}
\textcolor{black}{
It is out of the scope of our study to determine which benchmark is quantitatively and qualitatively best in evaluating QA systems. However, given the previous fine-grained properties of queries in the discussed benchmarks, and until further research on benchmarking QA over KGs is done, we can give insights on what the user can expect when using the discussed benchmarks in QA evaluation. In this section, and without loss of generality, we discuss evaluating QA on the two most common KGs in the literature (DBpedia and Freebase), excluding simple-factoid questions.\\
\textbf{QA over DBpedia:} \emph{QALD-1} to \emph{QALD-9}, and \emph{LC-QuAD} target DBpedia with a mix of simple and complex queries. \ul{If the user intends to evaluate their QA system over DBpedia, we recommend using all benchmarks from \emph{QALD} in addition to \emph{LC-QuAD}}, which is not a common practice in the literature. Based on the previously discussed properties, we notice that the \emph{QALD} benchmarks have better coverage of all possible different cases of using query keywords (Table~\ref{table:keywords}) and combination of operators (Table~\ref{table:operators}), while the \emph{LC-QuAD} queries exclusively use only the Distinct along with conjunctions of triple patterns. Also, the \emph{QALD} queries include a non-negligible number of queries that have more than three triple patterns (differs from one \emph{QALD} benchmark to another), whereas 100\% of the queries in \emph{LC-QuAD} have at most three triple patterns (Figure~\ref{fig:triples}). This is an indication that the \emph{QALD} benchmarks include more complex queries, which may prove more challenging for QA systems. This is demonstrated by the more complex query shapes in \emph{QALD} that do not exist in \emph{LC-QuAD} (i.e., Forest, Cycle, and Flower). However, since the \emph{QALD} benchmarks are manually created, they include 959 questions in total, whereas \emph{LC-QuAD} includes almost 5000 questions. This indicates better linguistic and vocabulary coverage in \emph{LC-QuAd}. Indeed, further investigations reveal that \emph{LC-QuAD}'s queries span 6 times more resources (entities) and 2 times more predicates than \emph{QALD}.
\\
\textbf{QA over Freebase:} The two benchmarks targeting Freebase that include queries are \emph{WebQuestions} and \emph{GraphQuestions}. \ul{We recommend using \emph{GraphQuestions} if the user intends to evaluate their QA system over Freebase}. While the differences are minor in terms of the operators used in the benchmarks (Table\ref{table:operators}), other syntactical and structural properties showcase the differences. \emph{GraphQuestions} include aggregate queries, while \emph{WebQuestions} do not (Table~\ref{table:keywords}). Also, \emph{WebQuestions} queries tend to be shorter (over 60\% of queries are single-triple-pattern and almost 100\% have no more than three triple patterns), whereas \emph{GraphQuestions} have longer and more complex queries (more than 35\% of queries have more than three triple patterns reaching 10 triple patterns) as shown in Figure~\ref{fig:triples}. This is further showcased by the percentage of simpler query shapes for both benchmarks, where most of the queries in \emph{WebQuestions} are chain queries, while \emph{GraphQuestions} have higher percentage of more complex shapes. \ul{\emph{WebQuestions} can be considered a better alternative to evaluate \textbf{simple questions} over Freebase} because it includes single-triple-pattern queries, while \emph{GraphQuestions} do not include such queries.}

\textcolor{black}{We will discuss our insights on benchmark selection with respect to the benchmarks that do not include queries in Section~\ref{sec:nl}.}



\section{Analysis of Natural Language Questions}
\label{sec:nl}
SPARQL queries are used to represent the graph patterns, whose matches are the desired answers in the KG. Therefore, analyzing SPARQL queries gives deeper insights into the structure of the questions in the benchmarks. However, not all benchmarks include SPARQL queries that correspond to the natural language questions (as previously shown in Table~\ref{table:datasets}). \textcolor{black}{In \system, we also analyze the natural language questions via the NLQ Analyzer to provide linguistic-based insights on the questions in the benchmarks. The analysis of the natural language questions in \system focuses on the type of the natural language questions (Section~\ref{subsec:typeLength}). This analysis gives the user insights on benchmark selections (Section~\ref{subsec:insights2}). In addition, to support the user with linguistic-based insights on the performance of the evaluated QA, \system provides the user with the most linguistically-similar questions to any question of interest. In order to support this feature, \system converts all natural language questions in the benchmarks to their vector space (Section~\ref{subsec:conversion}). This feature is demonstrated in Section~\ref{sec:eval}}.

\vspace{-2mm}
\subsection{Type of Questions}
\label{subsec:typeLength}
In \system, the questions are categorized into:
\squishlist
  \item Wh-questions: the type of questions that starts with a \emph{wh}-pronoun (i.e., \emph{What}, \emph{When}, \emph{Where}, \emph{Who}, \emph{Whom}, \emph{Which}, and \emph{Whose}). For example, \emph{\ul{Where} was the first ford motor company located?} (\emph{WebQuestions}) is a \emph{wh} question. This  excludes questions starting with the keyword \emph{Why}, which targets non-factoid questions that are out of scope for QA over KGs.
  \item How-questions: Questions starting with the keyword \emph{how} followed by \emph{many} or an adjective, such as \emph{\ul{How many} different currencies are used in the places governed by the president of France?} (\emph{LC-QuAD}) or \emph{\ul{How tall} is Michael Jordan?} (\emph{QALD-9}). Other non-factoid \emph{How} questions are not included.
  \item Yes/No questions: Questions that can be answered with \emph{yes} or \emph{no}: \emph{\ul{Is} Michelle Obama the wife of Barack Obama?} (\emph{QALD-9}).
  \item Requests: Direct requests, such as \emph{\ul{Can you name} all the states of the US?} (\emph{GraphQuestions}). \textcolor{black}{Such a question can be considered as a rephrasing of a \emph{What} question. However, we put these questions in a different category because they are usually handled differently in QA systems.}
  \item Topicalized questions: Questions in which an entity or prepositional phrase is topicalized for the purpose of emphasis: \emph{\ul{Adobe pdf} supports how many computing platforms?} (\emph{GraphQuestions}).
\squishend

\textcolor{black}{Table~\ref{type} shows the percentage of questions of each question type across all benchmarks. \textcolor{black}{We break the \emph{wh}-questions into their sub-types}. For \emph{wh}-questions, we also include many questions that start with a preposition followed by a \emph{wh}-word, such as \emph{\ul{in} which} or \emph{\ul{to} where}, because the single preposition can be easily moved to the end without changing the rest of the sentence. From these results, we observe that there is no dominant question type across the benchmarks and the distribution of query types vary widely.}



\small

\begin{table*}[]
\captionof{table}{Question frequency percentages (\%) by type for all benchmarks.}
          \label{type}
           \centering
\begin{tabular}{lrrrrrrrrrrrrrrrrrr}
\toprule
         & \multicolumn{1}{l}{{\rotatebox [origin=c]{70}{\textbf{QALD-1}}}} & \multicolumn{1}{l}{{\rotatebox [origin=c]{70}{\textbf{QALD-2}}}} & \multicolumn{1}{l}{{\rotatebox [origin=c]{70}{\textbf{QALD-3}}}} & \multicolumn{1}{l}{{\rotatebox [origin=c]{70}{\textbf{QALD-4}}}} & \multicolumn{1}{l}{{\rotatebox [origin=c]{70}{\textbf{QALD-5}}}} & \multicolumn{1}{l}{{\rotatebox [origin=c]{70}{\textbf{QALD-6}}}} & \multicolumn{1}{l}{{\rotatebox [origin=c]{70}{\textbf{QALD-7}}}} & \multicolumn{1}{l}{{\rotatebox [origin=c]{70}{\textbf{QALD-8}}}} & \multicolumn{1}{l}{{\rotatebox [origin=c]{70}{\textbf{QALD-9}}}} & \multicolumn{1}{l}{{\rotatebox [origin=c]{70}{\textbf{QALD}}}} & \multicolumn{1}{l}{{\rotatebox [origin=c]{70}{\textbf{LC-QuAD}}}} & \multicolumn{1}{l}{{\rotatebox [origin=c]{70}{\textbf{Web}}}} & \multicolumn{1}{l}{{\rotatebox [origin=c]{70}{\textbf{Graph}}}} & \multicolumn{1}{l}{{\rotatebox [origin=c]{70}{\textbf{Simple}}}} & \multicolumn{1}{l}{{\rotatebox [origin=c]{70}{\textbf{SimpleDB}}}} & \multicolumn{1}{l}{{\rotatebox [origin=c]{70}{\textbf{Temp}}}} & \multicolumn{1}{l}{{\rotatebox [origin=c]{70}{\textbf{Complex}}}} & \multicolumn{1}{l}{{\rotatebox [origin=c]{70}{\textbf{ComQA}}}} \\
         \midrule
What     & 6.53                                                             & 6.10                                                             & 7.30                                                             & 10.59                                                            & 11.08                                                            & 13.92                                                            & 11.70                                                            & 18.10                                                            & 13.48                                                            & 10.80                                                          & 53.44                                                             & 55.32                                                         & 33.08                                                           & 60.73                                                            & 57.19                                                              & 29.35                                                          & 32.00                                                             & 47.13                                                           \\
When     & 9.55                                                             & 6.40                                                             & 6.80                                                             & 3.12                                                             & 3.59                                                             & 4.18                                                             & 5.09                                                             & 4.76                                                             & 3.43                                                             & 6.00                                                           & 0.00                                                              & 4.12                                                          & 0.07                                                            & 0.01                                                             & 0.00                                                               & 22.03                                                          & 8.00                                                              & 10.66                                                           \\
Where    & 0.50                                                             & 0.58                                                             & 0.50                                                             & 0.93                                                             & 1.80                                                             & 2.32                                                             & 2.26                                                             & 2.54                                                             & 2.94                                                             & 1.88                                                           & 9.96                                                              & 18.57                                                         & 1.10                                                            & 7.37                                                             & 10.48                                                              & 4.48                                                           & 0.67                                                              & 4.19                                                            \\
Which    & 31.66                                                            & 26.74                                                            & 25.44                                                            & 35.83                                                            & 27.54                                                            & 25.99                                                            & 26.60                                                            & 20.95                                                            & 25.74                                                            & 27.25                                                          & 13.30                                                             & 1.81                                                          & 18.28                                                           & 13.20                                                            & 12.51                                                              & 9.44                                                           & 29.33                                                             & 6.96                                                            \\
Who      & 14.07                                                            & 14.24                                                            & 14.61                                                            & 12.15                                                            & 15.87                                                            & 16.71                                                            & 19.62                                                            & 19.05                                                            & 16.91                                                            & 15.68                                                          & 11.97                                                             & 19.82                                                         & 8.52                                                            & 11.52                                                            & 12.09                                                              & 33.52                                                          & 30.00                                                             & 21.27                                                           \\
Whom     & 0.00                                                             & 0.29                                                             & 0.76                                                             & 0.00                                                             & 0.00                                                             & 0.23                                                             & 0.19                                                             & 0.32                                                             & 0.00                                                             & 0.34                                                           & 0.12                                                              & 0.00                                                          & 0.17                                                            & 0.01                                                             & 0.03                                                               & 0.00                                                           & 0.00                                                              & 0.09                                                            \\
Whose    & 0.00                                                             & 0.00                                                             & 0.00                                                             & 0.00                                                             & 0.00                                                             & 0.00                                                             & 0.00                                                             & 0.00                                                             & 0.00                                                             & 0.00                                                           & 0.22                                                              & 0.00                                                          & 0.07                                                            & 0.06                                                             & 0.05                                                               & 0.00                                                           & 0.00                                                              & 0.04                                                            \\
How      & 10.55                                                            & 12.79                                                            & 13.35                                                            & 7.79                                                             & 10.48                                                            & 10.44                                                            & 11.51                                                            & 13.02                                                            & 10.78                                                            & 12.60                                                          & 1.26                                                              & 0.36                                                          & 9.27                                                            & 0.69                                                             & 0.41                                                               & 1.02                                                           & 0.00                                                              & 0.25                                                            \\
Yes/No   & 9.55                                                             & 9.88                                                             & 9.07                                                             & 7.48                                                             & 7.78                                                             & 6.96                                                             & 10.38                                                            & 10.48                                                            & 9.07                                                             & 7.63                                                           & 2.09                                                              & 0.00                                                          & 0.14                                                            & 1.20                                                             & 1.48                                                               & 0.00                                                           & 0.00                                                              & 0.01                                                            \\
Requests & 17.59                                                            & 22.67                                                            & 21.91                                                            & 21.18                                                            & 21.26                                                            & 18.79                                                            & 11.13                                                            & 10.79                                                            & 17.65                                                            & 16.88                                                          & 5.63                                                              & 0.00                                                          & 9.92                                                            & 3.31                                                             & 3.99                                                               & 0.00                                                           & 0.00                                                              & 0.98                                                            \\
Topical  & 0.00                                                             & 0.29                                                             & 0.25                                                             & 0.93                                                             & 0.60                                                             & 0.46                                                             & 1.51                                                             & 0.00                                                             & 0.00                                                             & 0.94                                                           & 2.01                                                              & 0.00                                                          & 19.38                                                           & 1.90                                                             & 1.77                                                               & 0.16                                                           & 0.00                                                              & 8.42                                                  \\
\bottomrule
\end{tabular}
\end{table*}

\normalsize

\vspace{-1mm}
\subsection{\textcolor{black}{Insights on Benchmark Selection (2)}}
\label{subsec:insights2}
\textcolor{black}{In Section~\ref{subsec:insights1}, we discussed our insights on benchmark selection based on our analysis of benchmarks that include queries. In this section, we enrich this discussion on these benchmarks and add new insights on the benchmarks that do not include queries. Again, and without loss of generality, we discuss evaluating QA on the two most common KGs in the literature (DBpedia and Freebase), excluding simple-factoid questions.\\
\textbf{QA over DBpedia:} We previously mentioned that both \emph{QALD} and \emph{LC-QuAD} need to be used to evaluate a QA system over DBpedia. This recommendation is solidified by our observations in this section. Both benchmarks vary in terms of the types of questions, where each benchmark covers the types that are not covered by the other. For example, \emph{LC-QuAD} does not include \emph{When}-questions but \emph{QALD} does, and \emph{QALD} does not include \emph{Whose}-questions but \emph{LC-QuAD} does.\\
\textbf{QA over Freebase:} Among the benchmarks that target Freebase, \emph{GraphQuestions} is the most comprehensive being the only benchmark that covers all question types. In contrast, the other benchmarks (\emph{WebQuestions}, \emph{TempQuestions}, \emph{ComQA}, and \emph{ComplexQuestions}) miss between 4 and 6 types. Another advantage of using \emph{GraphQuestions} is that it includes different utterances of the same question, which further challenges a QA system. For example, ``What celebrities are imitated by Will Ferrell?'', ``Who did Will Ferrell impersonate?'', etc. It is interesting to notice that \emph{GraphQuestions} include yes/no questions. However, the queries and answers do not reflect this type of questions by not using the SPARQL's \emph{ASK} keyword and having non-binary answers. For example, the question "\textit{Are there any digital cameras with the maximum aperture reaching 5.2?}" is answered by a list of these cameras instead of yes/no answer. Prior to \system, the user needed to adjust for this scenario by changing the answer in the benchmark to either Yes or No. \system handles this scenario for the user. \emph{TempQuestions} focuses on questions that include temporal aspects. So, it can be used as an auxiliary benchmark to another main benchmark (e.g., \emph{GraphQuestions}). The number of questions in \emph{ComplexQuestions} (150) and the number of types not included (6) are the main disadvantages of using it.}

\vspace{-1mm}
\subsection{Linguistically-Similar Questions}
\label{subsec:conversion}

Each token in the question is assigned a tag based on both the token itself and its context in the question. This process is known as Part-of-Speech (PoS) tagging. There have been numerous works in the literature on PoS tagging \cite{choi-etal-2015-depends}. State-of-the-art approaches use a bidirectional recurrent neural network (BiLSTMs) and a subsequent conditional random field (CRF) decoding layer in combination with word embeddings that are trained over a large corpus of text~\cite{akbik2018,ma2016}. In \system, we utilize a pre-trained model \cite{honnibal-johnson:2015} to annotate each token in a question with its predicted PoS tag. However, \system can be used with any PoS tagger. We define the \emph{tag dictionary} $td = \{t_{1}, t_{2}, \hdots, t_{n}\}$ as the set of all possible tags (of size $n$) that can be assigned to a question token. 
We use two types of tags: (1) The Universal POS (UPOS)\footnote{\url{https://universaldependencies.org/u/pos/}}, and (2) the more comprehensive Penn Treebank tag set \cite{Weischedel_2013}. The simple UPOS tags are the common representations for the word types, such as \emph{NOUN}, \emph{VERB}, and \emph{ADJ}, and do not code for any morphological features (i.e. the structure of words such as stems, root words, prefixes, and suffixes). On the other hand, the Penn Treebank tags are more specific. For example, the \emph{VERB} tag is further divided into \emph{MD} - auxiliary, \emph{VB} - base form, \emph{VBD} - past tense, \emph{VBG} - present participle, \emph{VBN} - past participle, \emph{VBP} - non-3rd person singular present, and \emph{VBZ} - 3rd person singular present. 
We denote the tag dictionary for UPOS as $td_{UPOS}$, where  $|td_{UPOS}| = 17$, and the tag dictionary for Penn Treebank as $td_{Penn}$, where $|td_{Penn}| = 36$. In section~\ref{sec:eval}, we qualitatively evaluate both choices of tag types in \system.

\setlength{\textfloatsep}{2pt}
\begin{algorithm}[t]
\SetKwData{Left}{left}\SetKwData{This}{this}\SetKwData{Up}{up}
\SetKwFunction{Union}{Union}\SetKwFunction{FindCompress}{FindCompress}
\SetKwInOut{Input}{input}\SetKwInOut{Output}{output}

\Input{\textcolor{black}{List} of tokens that represent the natural language question $tokens$, tag dictionary $td$, PoS tagger $tagger$}
\Output{Fixed-size vector representation $pos\_freq[|td|]$}
\BlankLine

Initialize $pos\_tag =$ [$i$] = 0\;
\For{$token$ \textbf{\upshape in} $tokens$}{
$pos\_freq\left[indexOf(tagger(token))\right]$++\;
}
Return $pos\_freq$\;

\caption{Vectorization of natural language questions using PoS tagging}
\label{alg:vectorization}

\end{algorithm}

Due to the richness of natural language, it is not expected that a sequence of PoS tags will reoccur frequently in the limited set of questions in the benchmarks. Therefore, for this linguistic feature (PoS tags), we use a function $f_{tag}$ that embeds the previously defined natural language representation (\textcolor{black}{list} of tokens) into a fixed-size vector space whose number of dimensions $n$ is equal to the size of the tag dictionary used. Algorithm~\ref{alg:vectorization} illustrates this function. The function uses the PoS tagger to tag each token in the question, then updates the frequency of the tag in the final vector representation. This representation not only captures the definition and the context of each token but also indirectly captures the length of the question by counting the frequencies of the occurrences of the tags.





We use the obtained vector representation to calculate how \emph{syntactically similar} a pair of questions is. In \system, we use the euclidean distance measure to represent how dissimilar two questions are. Formally, the distance between two vector representations of questions $p$ and $q$ is calculated as $d(p, q) = \sqrt{\sum_{i = 1}^n (p_i - q_i)^2}$, where $n = |td|$. 
This definition of distance is used to find the $k$ most syntactically close questions to a question of interest (e.g., a question that is incorrectly answered). The $k$ questions are shown to the user marked with whether they are correctly or incorrectly answered. To quickly find the $k$ nearest questions to a question of interest, we pre-compute a distance matrix between each pair of questions in the benchmark. This one-time process takes less than a minute.

It is worth noting that in addition to PoS tagging, we also implemented a similar approach based on dependency parse trees of the questions in the benchmark to map the questions to their corresponding vectorized representations before calculating the pair-wise euclidean distances. We did not observe major changes in the output (discussed in Section~\ref{sec:eval}). 

\newcommand\notsotiny{\@setfontsize\notsotiny\@vipt\@viipt}

\section{Evaluation}
\label{sec:eval}
\textcolor{black}{In this section, we experimentally evaluate six QA systems over different benchmarks using \system. We show how the variations in the previously discussed properties of the questions/queries affect the evaluation of QA systems (Section~\ref{subsec:qaeval}). We then discuss our insights on evaluating the QA systems using \system (Section~\ref{subsec:evalinsights}). We also use \system to perform a fine-grained analysis of the questions processed by the QA systems (Section~\ref{subsec:fineeval}). Finally, we qualitatively evaluate the different settings for highlighting linguistically-close questions to a question of interest (Section~\ref{subsec:linguisticeval}).}

\noindent\textbf{Evaluation Environment}: 
All experiments were run on a machine having an Intel(R) Core(TM) i9-9900K CPU@3.60GHz and 16 MB cache, 32 GB RAM, running on Ubuntu 18.04.

\noindent\textcolor{black}{\textbf{QA Systems}: We evaluate six QA systems \cite{diefenbach2017wdaqua,hu2017answering, singh2018reinvent, liang2021querying, dubey2016asknow, tanon2018platypus}. Three systems are remotely located and accessed via web services and three systems are run locally on the same computer as CBench. All systems are evaluated over the most recent versions of DBpedia and Wikidata.}

\squishlist
\item \noindent\textcolor{black}{ \textbf{WDAqua~\cite{diefenbach2017wdaqua}:} WDAqua is accessed remotely\footnote{\url{http://qanswer-frontend.univ-st-etienne.fr/}} and targets DBpedia and Wikidata. It ignores the syntax of the questions and focuses only on the semantics of the words in the question.}

\item \noindent\textcolor{black}{ \textbf{gAnswer~\cite{zou2014natural, hu2017answering}:}
gAnswer is accessed remotely\footnote{\url{http://ganswer.gstore-pku.com/}} and targets DBpedia only. It is based on finding a semantic graph to represent the question's intention and then maps the problem to a subgraph matching problem over the KG.}

\item \noindent\textcolor{black}{ \textbf{Qanary~\cite{singh2018reinvent,singh2018frankenstein}:}
Qanary is a modular QA framework that breaks the QA pipeline into independent modules that can be easily replaced to improve the overall quality of QA. Although there are many open-source modules are shared\footnote{\url{https://github.com/WDAqua/Qanary}}, most of them are not functioning. We selected three working components for this experiment which are NED-tagme (for the entity recognition module), Diambiguation-Property-OKBQA (for the relation mapping module) and Query Builder (for the query generation module). This QA pipeline is evaluated locally over DBpedia.}

\item \noindent\textcolor{black}{ \textbf{QAsparql~\cite{liang2021querying}:}
QAsparql uses machine learning to classify the question into one of three types: list, count or Boolean questions. After that, it detects entities and relations and builds the queries based on the question type. The generated queries are then ranked by a module based on Tree-LSTM~\cite{tai2015improved} to get the final query that is used to retrieve the system answer. QAsparql is oped-sourced\footnote{\url{https://github.com/Sylvia-Liang/QAsparql}} and is evaluated locally over DBpedia.
}

\item \noindent\textcolor{black}{ \textbf{AskNow~\cite{dubey2016asknow}:}
AskNow first normalizes the NLQ into an intermediary canonical syntactic form, named Normalized Query Structure, and then renders it into SPARQL queries. This QA system is open-sourced\footnote{\url{https://github.com/AskNowQA/AskNowNQS}}  and is evaluated locally over DBpedia.}

\item \noindent\textcolor{black}{ \textbf{AskPlatypus \cite{tanon2018platypus}:}
AskPlatypus is a multilingual QA system over Wikidata that is mainly based on query templates and slot filling. It is evaluated remotely\footnote{\url{https://askplatyp.us}} using questions targeting Wikidata from \emph{QALD-7} and \emph{QALD-8}.
}

\squishend

\noindent\textbf{Benchmarks}: The benchmarks included in \system are shown in Table~\ref{table:datasets}. For the fine-grained query analysis, we exclude the benchmarks that do not include SPARQL queries.

\noindent\textbf{Evaluation Metrics}: We evaluate the QA system using three metrics: Micro, macro, and global F-1 scores. First, we define precision, recall, and F-1 for an individual question $q_{i}$. If we assume that the set of answers in the gold standard for $q_{i}$ (correct answers) is $G_{q_{i}}$, and the set of answers from the QA system for this question is $A_{q_{i}}$, then we define the precision as $P_{q_{i}} = \frac{|G_{q_{i}} \bigcap A_{q_{i}}|}{|A_{q_{i}}|}$, recall as $R_{q_{i}} = \frac{|G_{q_{i}} \bigcap A_{q_{i}}|}{|G_{q_{i}}|}$, and F-1 as $F_{q_{i}} = \frac{2 P_{q_{i}} R_{q_{i}}}{P_{q_{i}} + R_{q_{i}}}$. If there are $n$ questions in the benchmark, then micro F-1 is calculated as follows: $P_{\mu} = \frac{\sum_{i=1}^{n}|G_{i} \bigcap A_{i}|}{\sum_{i=1}^{n}|A_{i}|}$, $R_{\mu} = \frac{\sum_{i=1}^{n}|G_{i} \bigcap A_{i}|}{\sum_{i=1}^{n}|G_{i}|}$, and $F_{\mu} = \frac{2 P_{\mu} R_{\mu}}{P_{\mu} + R_{\mu}}$. The macro F-1 is defined by calculating the F-1 scores per question and averaging the values for all questions. Formally, $F_{\Sigma} = \frac{\sum_{i=1}^{n} F_{q_{i}}}{n}$.
The global F-1 score evaluates the overall quality of the QA system in terms of the question it can answer correctly. If we assume that the set of questions in a benchmark is $Q$, the set of questions that are processed by the QA system (non-empty answers) is $S$, and the set of questions that is answered correctly is $C$, then we define the global precision as $P_{G} = \frac{|C|}{|S|}$, global recall as $R_{G} = \frac{|C|}{|Q|}$, and global F-1 as $F_{G} = \frac{2 P_{G} R_{G}}{P_{G} + R_{G}}$. The latter definition of the global scores is strict because a question is considered to be answered correctly only if $F_{q_{i}} = 1$. Therefore, in the literature, the notion of what is considered a correctly answered question is relaxed, which is referred to as \emph{partially correct answered questions}. A question is considered partially correctly answered if $0 < F_{q_{i}} < 1$. In this paper, we also study the effect of changing the $F_{q_{i}}$ value used to consider a question to be partially correctly answered on the reported $F_{G}$.

\subsection{Evaluation of QA Systems}
\label{subsec:qaeval}

\begin{table*}[t]
\captionof{table}{Evaluation of QA Systems over benchmarks targeting DBpedia/Wikidata. Benchmarks annotated with $\star$ include questions that target Wikidata.}
\label{table:F1Scores}
\begin{tabular}{lrrrrrrrrrrrrrrrrrr}
\toprule

  & \multicolumn{3}{c}{} & \multicolumn{3}{c}{}   & \multicolumn{3}{c}{\textcolor{black}{Qanary\cite{singh2018reinvent,singh2018frankenstein}}}   & \multicolumn{3}{c}{}   &
  \multicolumn{3}{c}{}   &\multicolumn{3}{c}{}\\

 & \multicolumn{3}{c}{\textbf{WDAqua}\cite{diefenbach2017wdaqua}} & \multicolumn{3}{c}{\textbf{gAnswer}\cite{zou2014natural, hu2017answering}}   &
 \multicolumn{3}{c}{\textbf{\textcolor{black}{(TM+DP+QB)}}}   & \multicolumn{3}{c}{\textcolor{black}{\textbf{QAsparql}\cite{liang2021querying}}}   &
 \multicolumn{3}{c}{\textcolor{black}{\textbf{AskNow}\cite{dubey2016asknow}}}   &\multicolumn{3}{c}{\textcolor{black}{\textbf{AskPlatypus}\cite{tanon2018platypus}}}\\

 \cmidrule(lr){2-4}
 \cmidrule(lr){5-7}
  \cmidrule(lr){8-10}
  \cmidrule(lr){11-13}
  \cmidrule(lr){14-16}
  \cmidrule(lr){17-19}

 \textbf{Basis}
 &\multicolumn{1}{c}{\textit{F\textsubscript{G}}}   
 &\multicolumn{1}{c}{\textit{F\textsubscript{$\mu$}}} 
 &\multicolumn{1}{c}{\textit{F\textsubscript{$\Sigma$}}}           
 &\multicolumn{1}{c}{\textit{F\textsubscript{G}}}   
 &\multicolumn{1}{c}{\textit{F\textsubscript{$\mu$}}} 
 &\multicolumn{1}{c}{\textit{F\textsubscript{$\Sigma$}}}  
 
&\multicolumn{1}{c}{\textcolor{black}{\textit{F\textsubscript{G}}}}   
 &\multicolumn{1}{c}{\textcolor{black}{\textit{F\textsubscript{$\mu$}}}} 
 &\multicolumn{1}{c}{\textcolor{black}{\textit{F\textsubscript{$\Sigma$}}}} 
 
 &\multicolumn{1}{c}{\textcolor{black}{\textit{F\textsubscript{G}}}}   
 &\multicolumn{1}{c}{\textcolor{black}{\textit{F\textsubscript{$\mu$}}}} 
 &\multicolumn{1}{c}{\textcolor{black}{\textit{F\textsubscript{$\Sigma$}}}}
 
 &\multicolumn{1}{c}{\textcolor{black}{\textit{F\textsubscript{G}}}}   
 &\multicolumn{1}{c}{\textcolor{black}{\textit{F\textsubscript{$\mu$}}}} 
 &\multicolumn{1}{c}{\textcolor{black}{\textit{F\textsubscript{$\Sigma$}}}}
 
 &\multicolumn{1}{c}{\textcolor{black}{\textit{F\textsubscript{G}}}}   
 &\multicolumn{1}{c}{\textcolor{black}{\textit{F\textsubscript{$\mu$}}}} 
 &\multicolumn{1}{c}{\textcolor{black}{\textit{F\textsubscript{$\Sigma$}}}}

 \\ \hline
QALD-1 	&	 
0.31   	    &	 
0.27    	    &
0.14    	    &	
0.44    	    &
0.18    	    &	 
0.24    	    &

\textcolor{black}{0.00}    	        &
\textcolor{black}{0.00}    	        &
\textcolor{black}{0.00}    	    &
\textcolor{black}{0.02}    	        &
\textcolor{black}{$\approx$0.00}     	        &
\textcolor{black}{0.01}     	            	    &
\textcolor{black}{0.12}    	        &
\textcolor{black}{$\approx$0.00}
&
\textcolor{black}{0.07}               &
\textcolor{black}{-}    	        &
\textcolor{black}{-}    	        &
\textcolor{black}{-}   

 \\
QALD-2 	&	 
0.32    	    &	 
0.17    	    &
0.16    	    &	 
0.41    	    &
0.08    	    &	 
0.21    	    &
\textcolor{black}{0.00}    	        &
\textcolor{black}{0.00}    	        &
\textcolor{black}{0.00}    	    &
\textcolor{black}{0.03}   	        &
\textcolor{black}{$\approx$0.00}     	        &
\textcolor{black}{0.01}    	            	    &
\textcolor{black}{0.14}    	        &
\textcolor{black}{$\approx$0.00}
&
\textcolor{black}{0.10}               &
\textcolor{black}{-}    	        &
\textcolor{black}{-}    	        &
\textcolor{black}{-}     	 	            	            	    
 \\
 QALD-3 	&	 
0.21    	    &	 
0.23    	    &
0.11    	    &	 
0.28    	    &
0.11    	    &	 
0.16    	    &
\textcolor{black}{0.05}    	        &
\textcolor{black}{$\approx$0.00}
&
\textcolor{black}{0.02}    	    &
\textcolor{black}{0.12}    	        &
\textcolor{black}{0.01}    	        &
\textcolor{black}{0.06}    	            	    &
\textcolor{black}{0.19}    	        &
\textcolor{black}{$\approx$0.00}
&
\textcolor{black}{0.13}               &
\textcolor{black}{-}    	        &
\textcolor{black}{-}    	        &
\textcolor{black}{-}     	   	            	            	    
 \\
 QALD-4 	    &	 
0.21   	        &	 
0.17    	    &
0.12   	        &	 
0.30    	    &
0.13    	    &	 
0.16    	    &
\textcolor{black}{0.03}    	    &
\textcolor{black}{$\approx$0.00}
&
\textcolor{black}{0.01}    	    &
\textcolor{black}{0.16}    	    &
\textcolor{black}{0.02}    	    &
\textcolor{black}{0.08}    	    &
\textcolor{black}{0.13}    	        &
\textcolor{black}{0.05}    	        &
\textcolor{black}{0.08}  
& 
\textcolor{black}{-}    	        &
\textcolor{black}{-}    	        &
\textcolor{black}{-}               	   	            	            	    
 \\
 QALD-5 	&	 
0.31    	    &	 
0.19    	    &
0.18    	    &	 
0.36    	    &
0.10    	    &	 
0.20    	    &
\textcolor{black}{0.04}    	        &
\textcolor{black}{$\approx$0.00}
&
\textcolor{black}{0.02}    	    &
\textcolor{black}{0.23}    	        &
\textcolor{black}{0.01}    	        &
\textcolor{black}{0.12}    	            	    &
\textcolor{black}{0.29}    	        &
\textcolor{black}{0.11}    	        &
\textcolor{black}{0.09}               &
\textcolor{black}{-}    	        &
\textcolor{black}{-}    	        &
\textcolor{black}{-}     	  	           	            	    
 \\
QALD-6 	&	 
0.36    	    &	 
0.15    	    &
0.24    	    &	 
0.39    	    &
0.09    	    &	 
0.25    	    &
\textcolor{black}{0.05}    	        &
\textcolor{black}{$\approx$0.00}
&
\textcolor{black}{0.02}    	    &
\textcolor{black}{0.29}    	        &
\textcolor{black}{0.01}    	        &
\textcolor{black}{0.17}    	            	    &
\textcolor{black}{0.30}    	        &
\textcolor{black}{0.09}    	        &
\textcolor{black}{0.09}               &
\textcolor{black}{-}    	        &
\textcolor{black}{-}    	        &
\textcolor{black}{-}     	    	            	            	    
 \\
 QALD-7$\star$ 	&	 
0.39    	    &	 
0.19    	    &
0.29    	    &	 
-    	        &
-    	        &
-    	            	    &
\textcolor{black}{0.07}    	        &
\textcolor{black}{0.02}    	        &
\textcolor{black}{0.06}     	    &
\textcolor{black}{0.30}    	        &
\textcolor{black}{0.14}    	        &
\textcolor{black}{0.17}    	            	    &
\textcolor{black}{0.37}    	        &
\textcolor{black}{0.14}    	        &
\textcolor{black}{0.15}               &
\textcolor{black}{0.15}    	        &
\textcolor{black}{$\approx$0.00}    	        &
\textcolor{black}{0.08}    	    	            	        	    
 \\
 QALD-8$\star$ 	&	 
0.43    	    &	 
0.17    	    &
0.33    	    &	 
-    	        &
-    	        &
-    	            	    &
\textcolor{black}{0.09}    	        &
\textcolor{black}{0.01}    	        &
\textcolor{black}{0.04}     	    &
\textcolor{black}{0.46}    	        &
\textcolor{black}{0.12}    	        &
\textcolor{black}{0.30}    	            	    &
\textcolor{black}{0.33}    	        &
\textcolor{black}{0.10}    	        &
\textcolor{black}{0.13}               &
\textcolor{black}{0.11}    	        &
\textcolor{black}{$\approx$0.00}
&
\textcolor{black}{0.06}    	   	            	            	    
 \\
 QALD-9 	&	 
0.43    	    &	 
0.20    	    &
0.32    	    &	 
0.44    	    &
0.10    	    &	 
0.30    	    &
\textcolor{black}{0.08}    	        &
\textcolor{black}{$\approx$0.00}
&
\textcolor{black}{0.07}    	    &
\textcolor{black}{0.32}    	        &
\textcolor{black}{0.02}    	        &
\textcolor{black}{0.19}    	            	    &
\textcolor{black}{0.26}    	        &
\textcolor{black}{0.07}    	        &
\textcolor{black}{0.08}               &
\textcolor{black}{-}    	        &
\textcolor{black}{-}    	        &
\textcolor{black}{-}     	   	            	            	    
 \\\hline
 Mean & 
 0.33 & 
 0.19 & 
 0.21 & 
 0.36 & 
 0.12 & 
 0.20
     	    &
\textcolor{black}{0.05}    	        &
\textcolor{black}{$\approx$0.00}
&
\textcolor{black}{0.03}    	    &
\textcolor{black}{0.21}    	        &
\textcolor{black}{0.04}    	        &
\textcolor{black}{0.12}    	            	    &
\textcolor{black}{0.24}    	        &
\textcolor{black}{0.06}    	        &
\textcolor{black}{0.10}               &
\textcolor{black}{0.13}    	        &
\textcolor{black}{$\approx$0.00}
&
\textcolor{black}{0.07}      	   	            	        \\
 Std  & 
 0.08 & 
 0.04 & 
 0.09 & 
 0.06 & 
 0.04 & 
 0.04    	    &
\textcolor{black}{0.03}    	        &
\textcolor{black}{$\approx$0.00}
&
\textcolor{black}{0.03}    	    &
\textcolor{black}{0.15}    	        &
\textcolor{black}{0.05}    	        &
\textcolor{black}{0.09}    	            	    &
\textcolor{black}{0.09}    	        &
\textcolor{black}{0.05}    	        &
\textcolor{black}{0.03}               &
\textcolor{black}{0.03}    	        &
\textcolor{black}{$\approx$0.00}    	        &
\textcolor{black}{0.01}     	   	            	        
\\\hline
 LC-QuAD 	&	 
0.20    	    &	 
0.03    	    &
0.15    	    &	 
-    	        &
-    	        &
-    	    &
\textcolor{black}{0.02}    	        &
\textcolor{black}{0.01}
&
\textcolor{black}{0.01}    	    &
\textcolor{black}{0.46}    	        &
\textcolor{black}{0.14}    	        &
\textcolor{black}{0.34}    	            	    &
\textcolor{black}{0.16}    	        &
\textcolor{black}{0.01}
&
\textcolor{black}{0.11}                &
\textcolor{black}{-}    	        &
\textcolor{black}{-}    	        &
\textcolor{black}{-}    	  	            	            	    
 \\\hline
 Mean & 
 0.32 & 
 0.18 & 
 0.20 & 
 0.36 & 
 0.12 & 
 0.20
     	    &
\textcolor{black}{0.04}    	        &
\textcolor{black}{0.01}
&
\textcolor{black}{0.03}    	    &
\textcolor{black}{0.24}    	        &
\textcolor{black}{0.05}    	        &
\textcolor{black}{0.15}    	            	    &
\textcolor{black}{0.23}    	        &
\textcolor{black}{0.06}    	        &
\textcolor{black}{0.10}               &
\textcolor{black}{0.13}    	        &
\textcolor{black}{$\approx$0.00}
&
\textcolor{black}{0.07}     		            	        
\\
 Std  & 
 0.09 & 
 0.06 & 
 0.08 &
 0.06 &
 0.04 & 
 0.04
     	    &
\textcolor{black}{0.03}    	        &
\textcolor{black}{0.01}
&
\textcolor{black}{0.02}    	    &
\textcolor{black}{0.16}    	        &
\textcolor{black}{0.06}    	        &
\textcolor{black}{0.11}    	            	    &
\textcolor{black}{0.09}    	        &
\textcolor{black}{0.05}    	        &
\textcolor{black}{0.03}               &
\textcolor{black}{0.03}    	        &
\textcolor{black}{$\approx$0.00}    	        &
\textcolor{black}{0.01}       	   	            	        
\\\hline
 
\bottomrule

\end{tabular}
\end{table*}
\normalsize

\begin{table}[t]
\captionof{table}{Evaluation of QA Systems over benchmarks targeting Freebase. ComplexQuestion is annotated with $\star$ because it includes 150 questions only.}
\label{table:F1Scores3}
\begin{tabular}{lrrr}
\toprule
 & \multicolumn{3}{c}{\textbf{WDAqua }\cite{diefenbach2017wdaqua}}\\ 
& \multicolumn{3}{c}{\textbf{(200 Questions)}}\\
 \cmidrule(lr){2-4}

 \textbf{Benchmark}
 &\multicolumn{1}{c}{\textit{F\textsubscript{G}}}   
 &\multicolumn{1}{c}{\textit{F\textsubscript{$\mu$}}} 
 &\multicolumn{1}{c}{\textit{F\textsubscript{$\Sigma$}}}

 \\ \hline
 GraphQuestions 	&	 
$\approx$0.00    	    &
$\approx$0.00    	    &
$\approx$0.00    	     
   	    
 \\
 WebQuestions 	&	 
0.22    	    &	 
0.12    	    &
0.12    	 
\\
 TempQuestions 	&	 
0.13    	    &	 
0.05    	    &
0.06    	    
 \\
 ComQA 	&	 
0.16    	    &	 
0.02    	    &
0.07    	    
 \\
 ComplexQuestions$\star$ 	&	 
0.32    	    &	 
0.11    	    &
0.19    	    
\\\hline
Mean & 0.17 & 0.06 & 0.09 \\
Std  & 0.12 & 0.05 & 0.07\\
\hline
\bottomrule

\end{tabular}
\end{table}
\normalsize



Table~\ref{table:F1Scores} shows the \textcolor{black}{six} systems and their scores for the 3 metrics used in \system with $F_{q_{i}} > 0$ for considering a question to be considered partially correctly answered. The effect of changing the value of the $F_{q_{i}}$ threshold will be discussed later in this section. gAnswer experiments on \emph{QALD-7} and \emph{QALD-8} are excluded since they include questions that target Wikidata, which is not supported by gAnswer. \textcolor{black}{The table shows that although the \emph{QALD} benchmarks come from the same source, the six systems have considerably large variations in their scores. 
We calculate the average and standard deviation for all scores. The standard deviation values range from 0.0 to 0.15, which are considered to be high for variations in F-1 scores. When adding the \emph{LC-QuAD} scores to the results, we notice a slight increase in the standard deviation of the global and micro scores and a slight decrease in the macro score. We will address the differences between the F-1 scores in Section~\ref{subsec:evalinsights}} 

Even though Freebase is deprecated since 2015, several benchmarks still target evaluating question answering over it. The following experiment is inspired by the recent trend of migrating Freebase to Wikidata~\cite{pellissier2016freebase}. We evaluate WDAqua using the benchmarks that originally target Freebase, but over Wikidata. The challenge to do this experiment is to guarantee that a chosen question from any of the benchmarks can be answered using Wikidata. We randomly sample questions from all the benchmarks and manually examine if the question can be answered using Wikidata by identifying the entities in the questions and the answers in the benchmark in Wikidata and finding a path that connects them, which resembles how a query should be written to retrieve the answers. We continue sampling until we find 200 Wikidata-answerable questions for each benchmark (with the exception of \emph{ComplexQuestions}, which includes only 150 questions). The results for this experiment are shown in Table~\ref{table:F1Scores3}. WDAqua performs considerably better for questions from \emph{WebQuestions} and \emph{ComplexQuestions} compared to the other benchmarks. This especially affects the standard deviation of $F_{G}$ (0.12), which is significant for F-1 scores.


\begin{figure}[t]
  \includegraphics[width=0.95\linewidth]{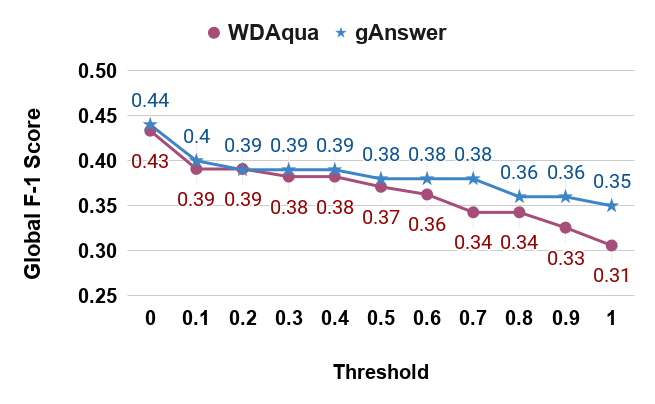}
  \caption{Evaluation of QA systems over \emph{QALD-9} using different thresholds for the $F_{q_{i}}$.}
  \label{fig:threshold}
\end{figure}

We also investigate the effects of using different thresholds for the $F_{q_{i}}$ for a question to be considered partially correct. For this analysis, we evaluate WDAqua and gAnswer over \emph{QALD-9} (the benchmark both systems perform best in). Figure~\ref{fig:threshold} shows that the two systems are significantly affected by the change of the $F_{q_{i}}$ value used. Although the two systems start at nearly the same value for $F_{G}$, where a question is considered partially correct if $F_{q_{i}} > 0$, WDAqua is more sensitive to stricter thresholds. Overall, there is a drop-in $F_{G}$ of 0.12 in WDAqua, where as gAnswer' score drops by 0.09. Both drops are considered significant for F-1 scores, in general. The drop in WDAqua's score is higher than that of gAnswer because WDAqua produces less-quality answers for individual questions compared to gAnswer. Therefore, a more restrictive quality threshold results in such a drop. It is also noticeable that the first step (from $F_{q_{i}} \geq 0$ to $F_{q_{i}} \geq 0.1$) has the most significant impact as $F_{G}$ drops by 0.04 for both systems, which is considered significant given the slightly stricter threshold.


\vspace{-1mm}
\subsection{\textcolor{black}{Insights on Evaluation Metrics and Benchmarks}}
\label{subsec:evalinsights}
\textcolor{black}{\system includes the three evaluation metrics that have been used in the literature: Micro F-1 ($F_{\mu}$), Macro F-1 ($F_{\Sigma})$, and Global F-1 ($F_{G}$). It is interesting to see that previous works on QA over KG usually use only a subset of the three metrics. It is also common to use only the Global F-1. Our evaluation suggests that \ul{using a subset of the metrics is misleading in evaluating QA systems}. Table~\ref{table:F1Scores} shows that the choice of a metric changes the order of the superior system. For example, if we consider the Global F-1 scores and use any of the \emph{QALD} benchmarks for evaluation, gAnswer is superior to WDAqua. However, if we consider the Micro F-1 scores, WDAqua is superior. This means that the quality of the answer of individual questions is better in WDAqua (better Micro F-1 scores). However, gAnswer is able to answer more questions correctly or partially correctly. Interestingly, the same observation applies for the same QA system in some cases. For example, when evaluating WDAqua using the benchmarks \emph{QALD-1} to \emph{QALD-5}, the Micro F-1 score is larger than the Macro F-1 score. However, using \emph{QALD-6} to \emph{QALD-9}, the Macro F-1 score is larger. This means that the questions WDAqua struggles with in \emph{QALD-6} to \emph{QALD-9} have long answers. This results in a worse hit on the micro scores than on the macro scores (average of scores of individual questions). The order also changes based on the benchmark used, suggesting that \ul{the choice of a set of benchmarks in evaluation can be misleading in evaluating QA systems}. For example, if we consider the Global F-1 scores and use \emph{QALD-7} to evaluate QAsparql and AskNow, we notice that AskNow is superior. However, if we use \emph{QALD-8}, a benchmark from the same authors but one year younger, QAsparql is superior. These variations pose questions on how the benchmarks are created. It is interesting to see that these variations do not exist only for the automatically-generated benchmarks, but also for benchmarks that are manually created. This leads us to conclude that \ul{more research effort needs to be put to better understand how to sufficiently represent a good balance between all the combinations of the properties that we proposed in our work to have good coverage of as many cases as possible that a QA system could face in practice}. Until such research is done, we recommend taking a comprehensive approach in evaluating QA systems by considering all evaluation metrics over the benchmarks that we recommended using in Sections~\ref{subsec:insights1} and~\ref{subsec:insights2}. Such a comprehensive approach is now facilitated by \system.}

\vspace{-1mm}
\subsection{Fine-Grained Evaluation}
\label{subsec:fineeval}
\begin{figure}[tbp]
  \includegraphics[width=0.93\linewidth]{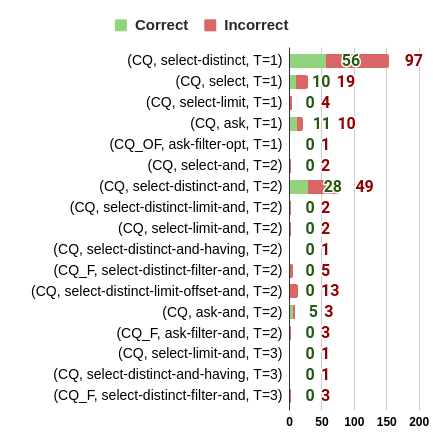}
  \caption{The properties of the \emph{Chain}-shaped queries of the queries from \emph{QALD-9} answered by WDAqua. \textcolor{black}{T represents the number of triple patterns}.}
  \label{fig:wdaChain}
\end{figure}

\begin{figure}[tbp]
  \includegraphics[width=0.98\linewidth]{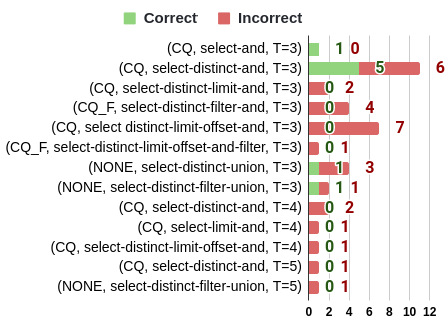}
  \caption{The properties of the \emph{Star}-shaped queries of the questions from \emph{QALD-9} answered by WDAqua. \textcolor{black}{T represents the number of triple patterns}}
  \label{fig:wdaStar}
\end{figure}

In light of the results in Section~\ref{subsec:qaeval}, we analyze the queries of the questions processed by the QA systems. Due to the lack of space, the experiments in this section focus on WDAqua, over \emph{Chain}-shaped and \emph{Star}-shaped queries of \emph{QALD-9}, using a threshold $F_{q_{i}} > 0.8$ to consider a question to be partially correct. We highlight the effectiveness of using \system's Debugging Mode to obtain the fine-grained results for these specific category of queries. Other settings (shapes, benchmarks, thresholds, and gAnswer) show similar results. It is also worth noting that these specific categories can be navigated to through the interactive report shown to the user if the user decides to evaluate the QA system using all the questions in a benchmark since we do a two-level grouping of all questions processed by the QA system when presenting them to the user. The first level of grouping is by shape, and the second level is by combinations of query properties. 

Figures~\ref{fig:wdaChain} and~\ref{fig:wdaStar} show the fine-grained analysis for the the aforementioned categories (\emph{Chain}-shaped and \emph{Star}-shaped queries, respectively). For each shape, each combination of properties shows how many questions were answered correctly or incorrectly by the QA system. Figure~\ref{fig:wdaChain} shows that there are 17 different combination of properties for all the \emph{Chain} queries in \emph{QALD-9}. Most of the \emph{Chain} queries are either conjunctive queries that use the keywords \emph{select} and \emph{distinct} and have one  \textcolor{black}{triple pattern} (56 questions are correctly answered and 97 are incorrectly answered), or conjunctive queries that use the same two keywords and have two  \textcolor{black}{triple patterns} (28 questions are correctly answered and 49 are incorrectly answered). The user can choose any of the categories in the figure to examine individual questions, in which case, they can use the natural language analysis features of \system (will be qualitatively evaluated in Section~\ref{subsec:linguisticeval}). It is worth noting that the \emph{Chain} shape subsumes the \emph{Single-Edge} shape, which is actually represented in this figure through the top five combinations of properties (i.e., T = 1).

Figure~\ref{fig:wdaStar} shows the \emph{Star} queries of \emph{QALD-9}. These are considered to be more complex queries, with at least three  \textcolor{black}{triple patterns}. The figure shows that WDAqua answered only 8 out of 38 questions whose query shape is \emph{Star}. It is also worth noting that the union of the results in the two figures is the expected results for the \emph{Tree}-shaped queries. The user can also navigate through the figures representing different shapes through the subsumption hierarchy of shapes.


\textcolor{black}{\textbf{Debugging the QA System Using Fine-Grained Analysis:} The main three components in any QA system are (1) Entity Recognizer, (2) Relation Mapper and the (3) Query Builder. For example, given the question \textit{What is the capital of Canada?} discussed in Table \ref{knn2}, a good QA system's Entity Recognizer would recognize the token \emph{Canada} as the corresponding resource (i.e. \emph{dbr:Canada} in DBpedia). The Relation Mapper maps the sequence of tokens \emph{is the capital of} to the corresponding predicate (i.e. \emph{dbo:capital} in DBpedia). Lastly, the Query Builder produces the following formal query that returns the answers from the targeted KG: \emph{SELECT ?x \{dbr:Canada dbo:capital ?x.\}}. These modules can be either dependent/independent on/of each other. The Query Builder mainly depends on a predefined/learned set of query templates as well as the entities and relations detected by the Entity Recognizer and the Relation Mapper. CBench can help the QA developers to learn more about the query templates that the QA does not handle using its fine-grained analysis. For example, Figure \ref{fig:wdaChain} shows that WDAqua fails to answer all the queries with properties like \emph{(CQ, select-distinct-limit-offset-and, T=2)} and \emph{($CQ_{F}$, select-distinct-filter-and, T=2)}. This indicates that the query builder of WDAqua may not be generating the query templates:
\emph{SELECT DISTINCT ?x\{ ... \} OFFSET m LIMIT n} and
\emph{SELECT DISTINCT ?x\{ ... FILTER(...)\}}. It is possible, however, that there is an issue with the Entity Recognizer and Relation Mapper for all the questions that were incorrectly answered. A more obvious example is when the fine-grained analysis shows that some questions whose queries have specific properties were answered correctly and some were not. The user can choose any of these questions for further investigations (discussed next).
}

\vspace{-1mm}
\subsection{Qualitative Evaluation of Linguistically-Similar Questions}
\label{subsec:linguisticeval}

\small
\begin{table}[t]
 \captionof{table}{Closest 5 question to the incorrectly answered question "What currency should I take to Dubai?", based on UPOS and Penn Treebank tag. The marks \cmark/\xmark denote correctly/incorrectly answered questions.}
         \label{knn1}
           \centering
        \setlength{\tabcolsep}{3pt} 
        \begin{tabular}{cl}
\toprule
	  \multicolumn{2}{c}{Linguistically Closest Questions - UPOS}  \\ \midrule
	  Answer &  \multicolumn{1}{c}{Question} \\ \midrule
	  \cmark & What currency should I take to Mauritius?  \\ 
	  \cmark & What currency should I take to Jamaica?  \\
	  \cmark & What currency should I take to Mexico?  \\ 
	  \cmark & What currency should you take to Morocco?  \\ 
	  \xmark & What prompted Shakespeare to write poetry?  \\ \midrule
	  \multicolumn{2}{c}{Linguistically Closest Questions - Penn Treebank}  \\ \midrule
	  Answer & \multicolumn{1}{c}{Question} \\ \midrule
	  \cmark & What currency should I take to Jamaica?  \\ 
	  \cmark & What currency should I take to Mauritius?  \\
	  \cmark & What currency should you take to Morocco?  \\ 
	  \cmark & What currency should I take to Mexico?  \\ 
	  \cmark & What kind of money should I take to Jamaica?  \\ \bottomrule
\end{tabular}

       \end{table}
\normalsize

\small
\begin{table}[t]
 \captionof{table}{Closest 5 question to the incorrectly answered question "What is the capital of Canada?", based on UPOS and Penn Treebank tag. The marks \cmark/\xmark denote correctly/incorrectly answered questions.}
          \label{knn2}
           \centering
        \setlength{\tabcolsep}{3pt} 
        \begin{tabular}{cl}
\toprule
	  \multicolumn{2}{c}{Linguistically Closest Questions - UPOS}  \\\midrule
	  Answer &  \multicolumn{1}{c}{Question} \\ \midrule
	  \xmark & What is the nickname of Edinburgh?  \\ 
	  \cmark & What is the capital of Cameroon?  \\
      \xmark & What is the meaning of Heydar ?  \\ 
	  \xmark & What is the currency of Chile?  \\ 
	  \xmark & What is the capital of Venezuela?  \\ \midrule
	  \multicolumn{2}{c}{Linguistically Closest Questions - Penn Treebank}  \\ \midrule
	  Answer &  \multicolumn{1}{c}{Question} \\ \midrule
	  \xmark & What is the title of Kakae?  \\ 
	  \cmark & What is the currency of Rhodesia?  \\
	  \cmark & What is the capital of Cameroon?  \\ 
	  \xmark & What is the capital of Venezuela?  \\ 
	  \xmark & What is the origin of Xynisteri?  \\ \bottomrule
\end{tabular}
        \end{table}
\normalsize

In this section, we assume that the user navigates through the fine-grained analysis figures and investigates a question of interest that is not correctly answered by the QA system. For benchmarks that do not include SPARQL queries, the user can directly choose any question from the benchmark.  \system finds the $k$ linguistically closest questions to this chosen question based on the approach discussed in Section~\ref{sec:nl}. This feature is especially beneficial in debugging the Entity Recognizer and the Relation Mapper components in the QA system. In this experiment, we use gAnswer as the QA system. Other configurations do not reflect major changes to the top-5 questions. Changing the value of the thresholds for $F_{q_{i}}$ to consider a question to be partially correctly answered results in possible changes in the label of the correctness of the question.

Table~\ref{knn1} shows the linguistically-closest 5 questions to the incorrectly answered question ``What currency should I take to Dubai?''. There is considerable overlap between the two tag dictionaries (4 out of 5 questions match). In both cases, we note that almost all of the similar questions are correctly answered, unlike the question of interest, possibly highlighting that the question is correctly answered if the entity is a country (one-hop connection to currency in the KG) rather than a city (two-hops connection to currency).

\textcolor{black}{Table~\ref{knn2} shows a different interesting scenario for the incorrectly answered question ``What is the capital of Canada?'', where for the top part (using UPOS), most of the similar questions are also incorrectly answered. We note that all the questions have the template ``What is the <place\_holder> of <place\_holder>?''. This indicates that the QA system struggles with this type of questions with only few exceptions. However, the question ``What is the capital of Cameroon?'' is answered correctly. This indicates that the QA system either could not identify \emph{Canada} as an entity, or \emph{capital of} as the predicate \emph{dbo:capital}. Indeed, investigating gAnswer reveals that the Relation Mapper (referred to in gAnswer as the Relation Recognition step) mistakenly mapped the relationship in the question to the \emph{type} predicate due to the sparsity of the relation mention dictionary that is built based on a large text corpus, which is considered an overfitting problem in learning the textual patterns.} 

\section{Discussion}
\label{sec:discussion}
In light of our work, we would like to highlight the following observations that we believe should be taken into consideration for future research work in the area of QA over KGs:
\squishlist
 \item Variations Among Benchmarks: Although all benchmarks target the same problem, there are high-degree variations with respect to several linguistic features in the natural language questions, and syntactical and structural features in the queries (Sections~\ref{sec:structuredAnalysis} and~\ref{sec:nl}). These variations pose questions on how the benchmarks are created. It is interesting to see that these variations do not exist only for the automatically-generated benchmarks, but also for benchmarks that are manually created. More research effort needs to be put to better understand how to sufficiently represent a good balance between all the combinations of the features that we proposed in our work to have good coverage of as many cases as possible that a QA system could face in practice.
 \item Correctness of Benchmarks: We noticed that the automatically-generated benchmarks may include some erroneous entries, which may affect the reported accuracy of the evaluated QA system. For example, it is interesting to notice that \emph{GraphQuestions} include yes/no questions. However, the queries and answers do not reflect this type of questions by not using the SPARQL's \emph{ASK} keyword and having non-binary answers. 
 \item Staleness: KGs are continuously evolving, while benchmarks are static by nature. Most of the KGs targeted by the benchmarks in \system now have more recent versions than those that were used to generate/create the benchmark. In our work, we proposed a simple solution by updating the benchmark entries based on the queried KG. However, this approach will not work if the KG's evolution touches its ontology or vocabulary. In fact, we noticed such behaviour in our benchmarks. Some queries no longer produce any results because the ontology of DBpedia changed. It is interesting to also notice that some other queries in earlier QALD benchmarks are no longer syntactically correct due to changes in the standard query language (SPARQL). This observation is especially concerning for benchmarks that do not include SPARQL queries, where it is not even possible to automatically update the answers in the benchmark and will require daunting human efforts to make the benchmark useful.
 \item KG-Related Issues: In our work, we encountered some interesting cases where the query in the benchmark is unnatural in the sense that it would have been written in a different way but for the representation of the answer in the KG. For example, the following query (from QALD-9) is associated with the question \emph{``What airlines are part of the SkyTeam alliance?''}
 \begin{verbatim}
SELECT DISTINCT ?uri WHERE { 
    ?uri a dbo:Airline { 
        ?uri dbo:alliance dbr:SkyTeam 
    } UNION { 
        ?uri dbo:Alliance dbr:SkyTeam 
    } 
}
\end{verbatim}
 Normally, the query should only include one graph pattern that consists of two  \textcolor{black}{triple patterns}. However, due to the co-existence of the two predicates in the query above (dbo:alliance versus dbo:Alliance), the query in the benchmark had to be written in that way. This issue in DBpedia was later corrected and the predicate dbo:Alliance was removed.
\squishend
\vspace{-3mm}
\section{Related Work}
\label{sec:related}
Retrieving data from databases using non-structured languages has been a topic of interest since the emergence of databases~\cite{baseball}. 
More recent works focused more on information retrieval (IR)-based QA systems, which were designed to handle free text. Then for the last two decades, many QA over KGs systems emerged due to the increasing amount of new structured data on the web that is in the form of KGs~\cite{Hoeffner2017}. Moreover, hybrid QA systems that retrieve data from both free text and KGs have also emerged (e.g. ~\cite{hawk,Beaumont2015}).
 
Due to the popularity of QA over KGs, several evaluation datasets emerged~~\cite{Unger2011,cabrio2012qakis,Cabrio2013,Unger2014,Unger2015,Unger2016,Usbeck2017,Usbeck2018,ngomo20189th,trivedi2017lc,Su2016,Berant2013,bordes2015large,Azmy2018,cai-yates-2013-large,jia2018tempquestions,Abujabal2017,abujabal-etal-2019-comqa} targeting multiple popular KGs like DBpedia~\cite{Auer2007}, Freebase~\cite{Bollacker2008}, Wikidata~\cite{vrandevcic2014wikidata}, and others. This variety of evaluation datasets (or benchmarks) poses a choice challenge when it comes to evaluating a new QA system. 

Closest to our work is \emph{GERBIL}~\cite{gerbil}, which allows uniform evaluation on diverse benchmarks. It focuses on question answering and semantic entity annotation. GERBIL is an extension of an earlier entity annotation system~\cite{usbeck2015evaluating}. With the new functionalities, GERBIL can be used to report micro and macro F-1 scores, along with execution time for multiple DBpedia-based benchmarks.

In \system, we support using any benchmark over any KG. \system also allows the user to add their own benchmarks over their own KG. In addition to providing the user with a report that includes single-numbered scores that indicates the overall performance with respect to the number of questions processed and correctly answered by the system, \system also provides a fine-grained report that groups processed questions per several syntactical and structural features of their corresponding SPARQL queries, and per the user request, highlights common characteristics or inconsistencies of the processed questions chosen by the user through contrasting them with the linguistically closest questions. This detailed information provided by \system helps the user identify strong and weak points of their QA system rather than just viewing single-number scores.

\section{Conclusion}
\label{sec:conclusion}
In this paper, we highlight the high-degree variations in the benchmarks used to evaluate question answering over knowledge graphs. We experimentally show that these variations affect the reported scores for multiple question answering systems. To overcome the effects of such variations, we introduce \system, a fine-grained benchmark suite that comes with a prepacked set of popular benchmarks that target multiple popular knowledge graphs. \system is easy to use and extensible. It does not only provide the user with the traditionally known quality scores, but it also gives a fine-grained analysis of the processed questions and their corresponding queries based on multiple features. This analysis can be used by the user to quickly identify the strong and weak points of their evaluated question answering system.

\bibliographystyle{abbrv}
\bibliography{All_Bib}

\end{document}